\documentclass{article}
\usepackage[preprint]{neurips_2026} 
\usepackage[utf8]{inputenc}
\usepackage[T1]{fontenc}
\usepackage{hyperref}
\usepackage{url}
\usepackage{booktabs}
\usepackage{amsfonts}
\usepackage{nicefrac}
\usepackage{microtype}
\usepackage{xcolor}
\usepackage{graphicx}
\usepackage{multirow}
\usepackage{subcaption}
\usepackage{amsmath}
\usepackage{orcidlink}
\usepackage{etoc}

\title{Decomposing the Generalization Gap in PROTAC Activity Prediction: \\ Variance Attribution and the Inter-Laboratory Ceiling}

\author{%
  Thor Klamt$^{1,*}$\,\orcidlink{0009-0005-6168-3655} \quad Wolfgang Nejdl$^{1,2}$\,\orcidlink{0000-0003-3374-2193} \quad Ming Tang$^{1,2}$\,\orcidlink{0000-0002-5993-5906} \\[0.6em]
  $^{1}$L3S Research Center, Leibniz Universit\"at Hannover, Appelstra\ss e 9a, 30167 Hannover, Germany \\
  $^{2}$Institute of Data Science (Knowledge-Based Systems),\\Faculty of Electrical Engineering and Computer Science,\\Leibniz Universit\"at Hannover, Appelstra\ss e 9a, 30167 Hannover, Germany \\[0.4em]
  $^*$\texttt{thor.klamt@gmail.com}
}

\begin{document}
\maketitle
\begin{abstract}
Machine-learning predictors of biochemical activity often exhibit large random-split-to-leave-one-target-out generalisation gaps that have been documented but not decomposed. We frame this as an evaluation-science question and use targeted protein degradation as the empirical test bed. PROTACs (proteolysis-targeting chimeras) are heterobifunctional small molecules that induce targeted protein degradation, with more than forty candidates currently in clinical trials; published predictors of PROTAC activity report AUROC values of $0.85$ to $0.91$ under random-split cross-validation, while the leave-one-target-out (LOTO) protocol of Ribes et al.\ reduces performance to approximately $0.67$. Random-split cross-validation rewards within-target interpolation, whereas LOTO measures the novel-target prediction that de-novo design depends on. We decompose this gap and identify inter-laboratory measurement variance as the dominant component, anchored by a within-target cross-laboratory cascade that bounds the inter-laboratory contribution at $0.124$ AUROC, substantially above the $0.05$ contribution from binarisation-threshold choice across four labelling schemes. Across eight published architectures and ESM-2 protein language models scaled to 3B parameters, LOTO AUROC plateaus near $0.67$ under the canonical evaluation protocol, with a comparable plateau under SMILES-level deduplication of the training partition; a 21-dimensional, 2000-trial hyperparameter optimisation across head architectures, molecular encoders, and protein encoders cannot break this ceiling, and the rank-1 single-seed configuration regresses by $0.161$ AUROC under multi-seed validation, matching a closed-form selection-bias prediction~\citep{bailey2014pseudomathematics,bailey2014deflated}. Few-shot $k{=}5$ stratified per-target retraining combined with ADMET features lifts 65-target LOTO AUROC from $0.668$ to $0.7050$, and post-hoc Platt scaling recovers raw output to within the $0.05$ well-calibrated threshold. We release PROTAC-Bench, a curated benchmark of 10,748 PROTAC degradation measurements across 173 targets and 65 LOTO folds, alongside the variance-decomposition framework, the per-target calibration protocol, and the evaluation code, providing a methodological template for variance attributions in other small-data therapeutic settings.
\end{abstract}

\section{Introduction}
\label{sec:intro}

PROTACs are heterobifunctional small molecules that bind a target protein and an E3 ubiquitin ligase to induce the target's polyubiquitination and proteasomal degradation. By eliminating the target rather than inhibiting its catalytic function, this mechanism in principle expands the druggable landscape to scaffold proteins, transcription factors, and other classes that lack the active-site geometry classical small-molecule inhibitors require, with more than forty PROTAC candidates currently in clinical trials and a medicinal-chemistry pipeline that depends on activity predictors for compound-library triage. Published PROTAC predictors~\cite{li2022deepprotacs,liu2025accurate,chen2025interpretable} report AUROCs of 0.85 to 0.91 under random-split cross-validation. \citet{ribes2024modeling} show that approximately 80 percent of random-split test entries share their UniProt target with a training entry, and that the leave-one-target-out (LOTO) protocol drops macro performance to approximately 0.67 against a matched random-CV macro of $0.755$ and pooled $0.902$. The pooled-to-macro aggregation thus separates from cross-target generalisation, with the latter dominated by cross-laboratory variance (Section~\ref{sec:decomposition}). The gap is not in dispute. What remains unanswered is whether it reflects models that cannot extrapolate to novel targets, or measurement variance that no binary-label predictor can resolve. Under the second, the gap reflects inter-laboratory measurement noise and labelling heterogeneity rather than learning failure, with cross-domain precedent in molecular-property and pair-input prediction~\cite{graber2025resolving,pahikkala2015toward}. Our analysis identifies inter-laboratory measurement variance as the dominant component, with the reproducibility floor as the operational ceiling at current dataset scale; the 65 LOTO-eligible targets reflect a deliberate trade-off in favour of within-target measurement density and multi-publication replicate structure over corpus size (Section~\ref{sec:benchmark:dataset}).

We report four contributions: (i) a variance-decomposition framework with three convergent ceiling estimates at 0.668 to 0.678 AUROC; (ii) an architecture-invariant performance plateau across eight published predictors and ESM-2 protein language models scaled to 3 billion parameters, with the rank-1 of 2,000 hyperparameter-optimisation trials regressing by 0.161 AUROC under multi-seed validation in agreement with the Bailey-L\'opez de Prado closed-form selection-bias prediction~\cite{bailey2014pseudomathematics,bailey2014deflated,cawley2010over}; (iii) a per-target recovery analysis where few-shot $k{=}5$ stratified retraining lifts 65-target LOTO AUROC from 0.668 to 0.7050, and a four-factor factorial decomposition that collapses an apparent $+0.042$ AUROC warhead-transfer effect to within seed noise once few-shot calibration is included as a co-factor; and (iv) PROTAC-Bench (10,748 measurements across 173 targets, 65 LOTO folds), evaluation code, Croissant metadata with MLCommons RAI extension, and a reusable dual-LLM metadata-enrichment pipeline.

\section{Related Work}
\label{sec:related}

\paragraph{PROTAC and targeted-degrader machine learning.}
Computational PROTAC activity prediction has been approached through deep architectures combining target-pocket encoders with PROTAC molecular encoders. DeepPROTACs~\cite{li2022deepprotacs} introduced a pocket graph convolutional network coupled to a BiLSTM over PROTAC fragment SMILES (0.847 random-split AUROC); DegradeMaster~\cite{liu2025accurate} extended this with an equivariant graph neural network over the ternary complex (approximately 0.94 random-split AUROC under supervised setting per their Table 2); PROTAC-STAN~\cite{chen2025interpretable} introduced ternary-attention over Morgan fingerprints, ESM-650M target embeddings, and an E3-ligase one-hot vector (0.883 random-split). Per-architecture replication notes appear in Appendices~\ref{app:degrademaster} and~\ref{app:negative}. \citet{ribes2024modeling} are the only prior authors to evaluate under LOTO, reporting an AUROC drop to 0.604; the random-CV-to-LOTO gap was documented but not decomposed. Generative approaches address PROTAC design rather than activity prediction~\cite{guan2023linkernet,li2024diffprotacs,li20233d} and assume a reliable activity oracle that the present work shows is currently unavailable. Ternary-complex structure prediction has advanced rapidly~\cite{dunlop2025predicting,riepenhausen2026ai,passaro2025boltz,xue2025se}; the structural approaches evaluated here predate AlphaFold-3 PROTAC accessibility, and Section~\ref{sec:collapse:negatives} demonstrates that structure-prediction quality under currently-evaluable tools does not translate to activity-prediction performance under held-out-target evaluation.\footnote{Concurrent published predictors and benchmarks not included in the canonical-evaluation sweep (AiPROTAC, DeepPROTACs 2.0, DeepPSA, Protap~\cite{yan2025protap}) require pipeline-specific reimplementation outside the canonical scope and are deferred to future work.}

\paragraph{Cold-split evaluation protocols.}
Distinguishing random-split from cold-split evaluation has cross-domain precedent. \citet{park2012flaws} introduced the C1, C2, C3, and C4 evaluation classes for protein-protein interaction prediction, formalised by \citet{pahikkala2015toward} for drug-target interaction prediction (PROTAC-Bench's LOTO protocol corresponds to the C3 setting), and \citet{bernett2024cracking} demonstrated that current sequence-based PPI predictors collapse to chance under leakage-free evaluation. Cold-split evaluation under scaffold or similarity-controlled protocols reveals performance degradation that random splits hide~\cite{liu2024welqrate,steshin2023hi,ektefaie2024evaluating,arevalo2024motive};the continuous similarity-resolution sweep in Appendix~\ref{app:spectra} integrates this framing with the binary LOTO protocol. \citet{tossou2024real} report up to 60 percent performance degradation under real-world molecular OOD; \citet{joeres2025data} introduce DataSAIL for principled split construction with quantitative leakage scoring; \citet{roberts2017cross} established block cross-validation for ecological models in the presence of spatial autocorrelation, the ecological analogue of inter-laboratory measurement variance.

\paragraph{Methodological lineage.}
The decomposition in Section~\ref{sec:decomposition} extends a methodological lineage of attributing apparent generalisation gaps to measurement-variance components: Cronbach generalizability theory~\citep{cronbach1972dependability}, pharmacogenomic and cross-source IC50 variance~\citep{haibe2013inconsistency,landrum2024combining}, train-test redundancy in molecular-property benchmarks~\citep{graber2025resolving}, hyperparameter-optimisation selection bias~\citep{cawley2010over,bailey2014pseudomathematics,bailey2014deflated,bouthillier2021accounting}, the transportability framework where LOTO is an explicit S-node intervention~\citep{pearl2014external}, architecture-invariance under cold-split evaluation~\citep{xia2023understanding,deng2023systematic,gulrajani2020search}, accidental-taxonomy and shortcut-learning interpretations of inverted PLM scaling~\citep{hallee2025protein,cheng2024training,geirhos2020shortcut,kumar2022fine}, simple-baseline outperformance of parameterised meta-learning~\citep{tian2020rethinking,chen2019closer,stanley2021fs}, selective-classification inversion under dataset shift~\citep{jones2020selective,ovadia2019can}, and broader generalizability concerns~\citep{yarkoni2022generalizability,d2022underspecification}. The present work contributes four methodological moves anchored to this lineage: variance-share decomposition of the cold-split gap, the Bailey-L\'opez de Prado closed-form selection-bias anchor applied to ML hyperparameter regression, within-target multi-publication replicate structure as a benchmark feature, and Platt scaling recovery under cold-target distributional shift.

\section{PROTAC-Bench}
\label{sec:benchmark}

\subsection{Dataset}
\label{sec:benchmark:dataset}

PROTAC-Bench is constructed from the publicly available PROTAC-DB~3.0~\cite{ge2024protacdb}, the curated benchmark of \citet{ribes2024modeling}, and DegradeMaster~\cite{liu2025accurate}, merged into 10,748 entries spanning 173 unique UniProt protein targets, with within-target measurement density and inter-laboratory replicate structure prioritised over corpus size since the variance attribution that anchors the analysis (Section~\ref{sec:decomposition}) requires per-target depth that catalogue-style compendia at larger compound counts cannot support. Each entry is associated with a SMILES representation of the PROTAC molecule, the target UniProt identifier, the recruited E3 ligase (CRBN, VHL, or other), continuous DC50 and Dmax measurements where available, and a binarised activity label following the convention of prior work~\cite{ribes2024modeling}: active if DC50 below 1\,$\mu$M OR Dmax above 50 percent. The release CSV carries binary activity labels; continuous DC$_{50}$ and D$_{\text{max}}$ values are pointed to upstream PROTAC-DB~3.0~\cite{ge2024protacdb} where source-traceable, with user-side recovery supported by the released compound-target identifier mapping. The activity rate across the full benchmark is 0.658. The LOTO-cohort positive rate of $0.672$ produces a random-baseline AUPRC equal to the prior (versus AUROC's random-baseline of $0.500$); AUPRC absolute values fall within $\pm 0.02$ of AUROC across architectures, but at AUROC $0.668$ the corresponding AUPRC sits at the prior and indicates no-skill performance above the base-rate predictor on the precision-recall surface despite above-chance ranking on AUROC. The two metric-relative-to-baseline interpretations should therefore be read independently rather than as parallel indicators of model quality, following \citet{mcdermott2024closer}; AUPRC values should be read against the $0.672$ base rate rather than the $0.500$ AUROC random-baseline. Per-architecture AUPRC values are reported in Appendix~\ref{app:robustness}. Eligibility filters defining the four evaluation cohorts (LOTO at 65 targets, LOFO at 61 targets, within-target cross-lab at 36 targets) are reported in Table~\ref{tab:dataset}. The E3-ligase distribution is heavily skewed (CRBN $n{=}7727$, VHL $n{=}2896$, other $n{=}125$), motivating the cross-E3 robustness analysis in Appendix~\ref{app:robustness}. A scaffold-split alternative was considered and rejected: 5-fold scaffold cross-validation reaches $0.897$ AUROC, statistically indistinguishable from random cross-validation at $0.902$. Under 5-fold scaffold-CV, $99.6$ percent of test compounds share their target with at least one training compound (vs $0$ percent under LOTO); $88$ of $173$ targets appear in all five scaffold folds. The $7{,}427$ unique Murcko scaffolds across $10{,}748$ compounds (Table~\ref{tab:dataset}) leave same-target compounds distributed across many distinct scaffolds, so a scaffold split does not isolate targets and the chemical-novelty shift that LOTO enforces is not reproduced. LOTO is therefore adopted as the canonical evaluation protocol throughout.

PROTAC-Bench is not in competition on raw size with the larger catalogue-style compendia: TPDdb~\cite{qin2026tpddb} (October 2025, 22,183 entries) and PROTAC-PatentDB~\cite{cai2025protac} (November 2025, 63,136 patent compounds)substantially exceed PROTAC-Bench in compound count but yield only 10 LOTO-evaluable targets in TPDdb under PROTAC-Bench's eligibility filters versus 65 in PROTAC-Bench, since most of their volume derives from patent enumeration without per-target activity depth (Appendix~\ref{app:robustness}, Table~\ref{tab:corpora}). PROTAC-Bench's distinguishing curation feature is within-target measurement density and multi-publication replicate structure: the 36-target cross-laboratory cohort with at least three publications per target enables the inter-laboratory variance attribution in Section~\ref{sec:decomposition}, and a reusable dual-LLM metadata-enrichment pipeline (Appendix~\ref{app:metadata}) recovers cell-line, readout-method, timepoint, and concentration annotations on 70.9 percent of the source corpus.

\begin{table}[!htb]
\centering
\caption{PROTAC-Bench dataset summary.}
\label{tab:dataset}
\small
\begin{tabular}{@{}ll@{}}
\toprule
Property & Value \\
\midrule
Total entries & 10,748 \\
Unique targets (UniProt) & 173 \\
LOTO-eligible targets ($n\geq10$, pos rate $\in[0.1,0.9]$) & 65 \\
LOFO-eligible targets (covered by 22-family map) & 61 \\
Cross-lab eligible targets ($\geq 3$ papers, $\geq 5$ entries each) & 36 \\
E3 ligase distribution & CRBN 7,727 / VHL 2,896 / other 125 \\
Activity rate (full benchmark / LOTO cohort) & 0.658 / 0.672 \\
Label criterion & DC50 $<$ 1\,$\mu$M OR Dmax $>$ 50\% \\
Unique Murcko scaffolds (singletons) & 7,427 (5,771) \\
\bottomrule
\end{tabular}
\end{table}

\subsection{Evaluation Protocol}
\label{sec:benchmark:protocol}

Four primary evaluation splits are defined, with random cross-validation referenced as the published-AUROC baseline only. LOTO removes all entries for one held-out UniProt target from training and evaluates the model on the held-out target. LOFO removes all entries for one held-out protein family, providing a stricter test of structural generalisation. Temporal-prospective evaluation trains on entries from publications before 2023 and tests on 2024 entries. Within-target cross-lab evaluation holds out one publication's entries within each multi-publication target, providing the inter-laboratory measurement-variance anchor for Section~\ref{sec:decomposition}. The canonical evaluation metric is macro-mean AUROC across held-out folds rather than pooled AUROC, since the held-out-target deployment regime is target-distributional rather than compound-distributional; the choice and its consequences for the gap decomposition are detailed in Section~\ref{sec:decomposition}. The split hierarchy under matched canonical RF settings is: random CV pooled at 0.902, scaffold CV at 0.897, LOTO macro-mean at 0.668, LOFO macro-mean at 0.616.

\subsection{Baseline and Released Artefacts}
\label{sec:benchmark:baseline}

The canonical baseline is RandomForestClassifier (n\_estimators=200, min\_samples\_leaf=3, class\_weight=balanced) applied to 2048-bit Morgan fingerprints at radius 2, evaluated across 10 canonical seeds \{7, 13, 29, 42, 43, 44, 53, 71, 89, 97\}. The 10-seed validated LOTO macro-mean AUROC is $0.668 \pm 0.005$ with 95 percent CI [0.664, 0.672]. Ten seeds were chosen because per-target heterogeneity (approximately 0.19 standard deviation across the 65-target cohort) dominates seed stochasticity (approximately 0.005). Fingerprint configuration is largely insensitive to choice: 15 radius/bit-width combinations yield results within 0.01 AUROC of canonical. The released artefacts comprise the dataset CSV, fold assignment files for the four primary evaluation splits (LOTO, LOFO, temporal-prospective, within-target cross-lab), the dual-LLM-enriched metadata file, evaluation code, and Croissant metadata at MLCommons Croissant 1.0 schema compliance with all twenty MLCommons RAI extension fields populated (validation detail in Appendix~\ref{app:reproduce}). The dataset is hosted at \url{https://huggingface.co/datasets/ThorKl/protac-bench} under CC-BY-4.0. The evaluation code and reproducibility scripts are hosted on GitHub at \url{https://github.com/ThorKlm/PROTAC-Bench} under MIT. A reusable dual-LLM metadata-enrichment pipeline is released alongside the dataset (70.9 percent coverage on the 9,384-row source corpus, full methodology in Appendix~\ref{app:metadata}); the \texttt{reproduce.sh} script regenerates canonical results (Appendix~\ref{app:reproduce}).

\section{Universal Architecture Plateau Across Eight Predictors}
\label{sec:collapse}

\subsection{Architecture-Invariant Performance Plateau}
\label{sec:collapse:architectures}

LOTO performance plateaus within a 0.13 AUROC band across all eight architectures we evaluated. The panel spans published PROTAC models (DeepPROTACs~\cite{li2022deepprotacs}, DegradeMaster~\cite{liu2025accurate}, the Ribes ProtTrans baseline~\cite{ribes2024modeling}) and standard chemical baselines (Random Forest with Morgan fingerprints, GIN, D-MPNN, ChemBERTa-2); PROTAC-STAN~\cite{chen2025interpretable} is included via partial replication of its ternary-attention component (Appendix~\ref{app:negative}, single-seed 0.718 regressing to 0.656 under 3-seed validation). The architecture-invariance pattern observed here differs from the analogous finding of \citet{xia2023understanding} under random-CV in being driven by the inter-laboratory measurement-variance ceiling identified in Section~\ref{sec:decomposition} rather than by representational adequacy of simple baselines. The largest matched-cohort delta (DegradeMaster on the 19-target intersection of its 27-target cohort with canonical LOTO eligibility, Table~\ref{tab:collapse}) is bounded above by +0.024 AUROC under matched 7-seed fast-protocol evaluation, statistically indistinguishable from the canonical baseline under Holm correction across the eight-architecture sweep (Appendix~\ref{app:degrademaster}) and within the 0.124 inter-laboratory variance band documented in Section~\ref{sec:decomposition}.\footnote{The eight-architecture panel is protein-encoder-centric; structure-prediction PROTAC variants (BoltzDesign-class, ProteinMPNN-PROTAC) predict ternary geometry rather than activity and are out of scope for the activity-prediction comparison reported here.}

Protein language model (PLM) scaling presents a non-monotonic relationship under LOTO evaluation, with ESM-2 peaking at the 150M parameter scale ($0.691$) and reduced performance at smaller (8M: $0.674$, 35M: $0.665$) and larger (650M: $0.667$, 3B: $0.656$) scales under maximal pretraining overlap (all 65 LOTO-eligible targets at 100 percent UniRef100 identity to the ESM-2 corpus). Larger PLMs do not improve LOTO under this configuration; the result is confounded by leakage and consistent with an accidental-taxonomist mechanism~\cite{hallee2025protein} in which larger PLMs identify target taxonomy more sharply on a panel they have already seen verbatim. The $0.691$ peak at 150M sits within the $0.124$ inter-laboratory variance band of Section~\ref{sec:decomposition}; the architecture-invariance claim is ceiling-bounded rather than pointwise. The methodologically appropriate test of PLM architectural saturation requires UniRef50-cluster-level holdout or masked-cluster training~\cite{hermann2024beware,szymborski2026flaw} and is deferred to a future revision; the architecture-invariance claim therefore scopes to PLM scale providing no benefit on this maximally-leaked panel rather than to PLM scale providing no benefit in general (full UniRef diagnostic in Appendix~\ref{app:plm}).

\begin{figure}[!htb]
  \centering
  \includegraphics[width=0.85\linewidth]{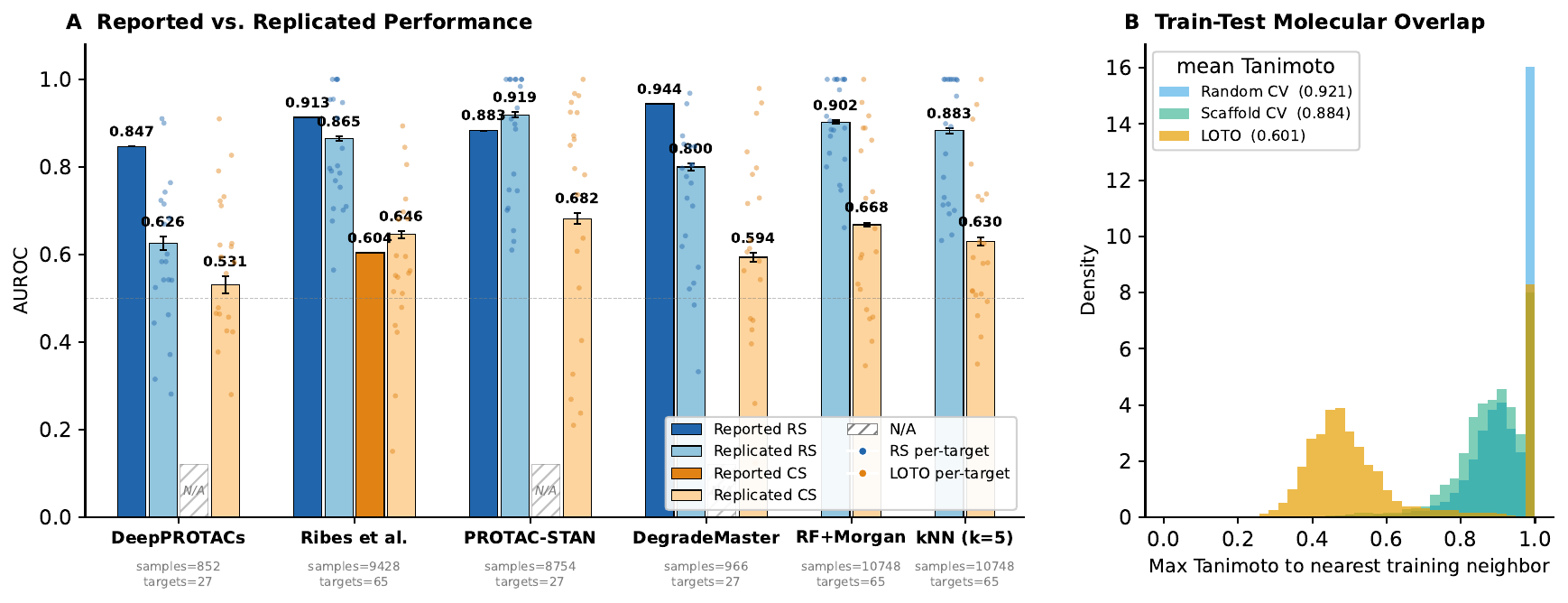}
  \caption{\textbf{The held-out-target performance gap and train-test molecular overlap.}
  (A)~Reported versus replicated AUROC across four published PROTAC predictors
  (DeepPROTACs, Ribes et al., PROTAC-STAN, DegradeMaster) and two reference
  baselines (RF+Morgan, kNN). Per-target AUROCs are overlaid as scatter dots on
  the replicated bars.
  (B)~Maximum Tanimoto similarity to nearest training-set neighbor, density
  across protocols. Random cross-validation yields a tight high-similarity
  distribution; scaffold cross-validation reduces overlap moderately; LOTO
  produces a broad low-similarity distribution.}
  \label{fig:collapse}
\end{figure}

\begin{table}[!htb]
\centering
\caption{Architecture invariance under canonical LOTO evaluation. Standard deviations across canonical seeds (3 to 10 depending on architecture). Per-target heterogeneity is approximately 0.19 across all rows. Extended table covering GIN, D-MPNN, and DeepPROTACs is reported in Appendix~\ref{app:degrademaster}. $^\dagger$ indicates values reported by original authors under their canonical random-split protocol rather than re-evaluated here. The RF+Morgan row reports $n_{\text{tgt}} = 64$ rather than 65 because one LOTO-eligible target failed per-seed evaluation under the matched canonical-RF settings used for the Table; the Section~\ref{sec:benchmark:baseline} canonical baseline of $0.668 \pm 0.005$ is computed across all 65 LOTO-eligible targets.}
\label{tab:collapse}
\footnotesize
\setlength{\tabcolsep}{4pt}
\begin{tabular}{@{}lccccc@{}}
\toprule
Method & RCV pool & RCV macro & LOTO macro & $\Delta_{\text{macro}}$ & $n_{\text{tgt}}$ \\
\midrule
RF + Morgan 2048 r2       & $0.902 \pm 0.001$ & $0.755 \pm 0.006$ & $0.668 \pm 0.005$ & $-0.087$ & 64 \\
ChemBERTa-2               & $0.875 \pm 0.002$ & $0.717 \pm 0.010$ & $0.663 \pm 0.008$ & $-0.054$ & 65 \\
Ribes (ProtTrans)         & $0.913 \pm 0.001$ & $0.716 \pm 0.007$ & $0.637 \pm 0.011$ & $-0.079$ & 65 \\
DegradeMaster (canonical) & $\sim 0.944^\dagger$ & $0.800^\dagger$  & $0.594 \pm 0.014$ & $-0.206$ & 61 \\
\bottomrule
\end{tabular}
\end{table}

\subsection{Hyperparameter Optimisation Cannot Break the Ceiling}
\label{sec:collapse:hpo}

We performed a 21-dimensional hyperparameter optimisation across 2,000 trials and find that the ceiling is not broken: the rank-1 single-seed objective of 0.764 regresses to $0.603 \pm 0.012$ under 5-seed validation, an inflation of 0.161 AUROC of the single-seed objective above the multi-seed-validated mean. Ranks 2 through 10, each validated under 5 canonical seeds, regress to the 0.659 to 0.679 range and are statistically indistinguishable from the canonical RF + Morgan baseline at 0.668 (paired Wilcoxon $p > 0.1$); the rank-1 regression illustrates the Cawley-Talbot selection-bias mechanism while ranks 2 through 10 converge at the architecture-invariance ceiling. Configurations selected by HPO under-represent warhead transfer at 4 percent of top-50 trials versus 36 percent base rate, indicating that single-evaluation HPO does not discover signals requiring multi-seed validation, consistent with \citet{bouthillier2021accounting}. The 0.161 regression matches the Bailey-L\'opez de Prado closed-form maximum-over-trials prediction $E[\max] \approx \sqrt{2 \log N}\,\sigma = 0.167$ at $N{=}2000$, $\sigma \approx 0.043$~\citep{bailey2014pseudomathematics,bailey2014deflated}, to within 4 percent under independent-trials assumption (full detail in Appendix~\ref{app:hpo}). A continuous SPECTRA-style similarity-resolution sweep~\cite{ektefaie2024evaluating} places LOTO at $0.668$ below the most restrictive measured similarity threshold ($s=0.50$, AUROC $0.652$; Appendix~\ref{app:spectra}).

\subsection{Additional Approaches Below the Ceiling}
\label{sec:collapse:negatives}

Beyond the eight tabulated architectures, PLM embeddings, EGNN encoders, Boltz-2 ternary structure features~\cite{passaro2025boltz}, and 22 metadata-feature variants yield LOTO AUROC within $0.01$ of the canonical baseline (Appendices~\ref{app:plm},~\ref{app:egnn},~\ref{app:geometric},~\ref{app:metadata_tasks}); the within-target cross-lab cascade $0.802 \to 0.678 \to 0.653$ anchors Section~\ref{sec:decomposition}, with EGNN pocket-shuffle and zero-pocket controls within $0.013$ AUROC of original-pocket performance (Appendix~\ref{app:egnn}).

\section{Variance-Share Decomposition: Inter-Laboratory Variance Sets Most of the Ceiling}
\label{sec:decomposition}

We decompose the gap, anchor each component empirically, and test the dominance ordering. The within-target cross-lab cascade attributes a 0.124 AUROC bound to inter-laboratory measurement effects (random-CV 0.802, cross-lab 0.678, LOTO 0.653 on the 36-target cohort, paired $n=35$), with a complementary target-clustered variance-share analysis assigning $\omega^2 = 0.256$ of cohort variance to the target facet under bias correction (95 percent CI $[0.110, 0.421]$, target heterogeneity inclusive of inter-laboratory variance). The unbiased $\omega^2$ values are reported as primary because eta-squared is upward-biased at small $n$~\cite{okada2013omega}; negative $\omega^2$ values, which appear for the laboratory-times-binarisation interaction term under the small-$n$ bias correction, indicate effects statistically indistinguishable from zero rather than numerical errors and should be read as zero (full ANOVA tabulation including eta-squared corroboration and interaction-term detail in Appendix~\ref{app:robustness}). The 0.124 inter-laboratory anchor itself is supported by a target-clustered bootstrap 95 percent CI of $[0.0512, 0.1971]$ and leave-one-target-out sensitivity range $[0.1132, 0.1308]$ across the 12 replicate-bearing targets (Appendix~\ref{app:robustness}).

The variance-share analysis is triangulated by four independent measurement-protocol bounds: the 0.124 AUROC inter-laboratory bound from the cross-lab cascade on the 36-target cohort (random-CV 0.802, cross-lab 0.678, LOTO 0.653; median 3.7-fold cross-lab DC50 change on identical compounds, Appendix~\ref{app:robustness}); the 0.05 binarisation spread across four labelling schemes; the 0.008 cross-DOI conflict-removal contribution; and the 0.02 to 0.03 residual LOTO distributional shift (Appendix~\ref{app:noise_calibration}). The cross-lab anchor dominates; remaining contributions sit within the small-$n$ bias correction or seed noise. The alternative inherent-difficulty interpretation is inconsistent with the SPECTRA sweep at $s=0.50$ (AUROC 0.652, Appendix~\ref{app:spectra}) and with the within-target cross-lab evaluation at 0.678 under removed structural shift. Cross-lab same-compound pairs are retained without deduplication because identical compounds measured across publications are the empirical signal anchoring the variance estimate (Appendix~\ref{app:robustness}, 28 cross-paper comparisons across 12 replicate-bearing targets, median 3.7-fold).

Three closely related views of the LOTO ceiling fall in the 0.668 to 0.678 range under similar aggregation choices, identifying inter-laboratory measurement variance as accounting for $0.117$ of the $0.149$ AUROC random-CV-to-LOTO gap on the 36-target cross-laboratory cohort (approximately 81 percent), and a comparable fraction of the smaller 65-target macro-gap of $0.087$ under family-composition transfer (Appendix~\ref{app:robustness}). The per-target standard deviation of LOTO AUROC across the 65-target cohort is $0.190$, reported separately in Appendix~\ref{app:lofo_per_family} as a per-target variance measure rather than as a between-protocol gap. The dominance argument rests on partially-overlapping bounds rather than a strict additive identity: the four anchors (0.124, 0.05, 0.008, 0.02 to 0.03) sum to approximately the 0.18 to 0.20 macro-mean gap as a coarse upper-bound triangulation, and the variance-to-AUROC translation is empirically anchored by the Appendix~\ref{app:noise_calibration} synthetic-noise calibration (0.124 bound projects to 21 to 27 percent equivalent flip rate across three noise models). The roughly 80 to 85 percent attribution is conditional on within-cohort identification. The decomposition predicts that per-target calibration with a small number of measured compounds recovers the inter-laboratory variance component; Section~\ref{sec:recovery} confirms this empirically.

\begin{figure}[!htb]
  \centering
  \includegraphics[width=0.45\linewidth]{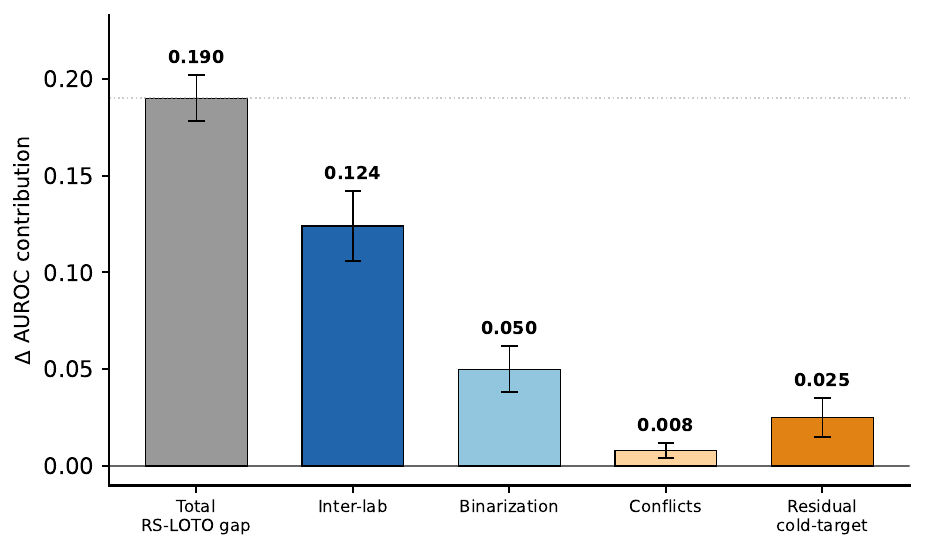}
  \caption{\textbf{Triangulating bounds on the random-CV-to-LOTO gap.} Methodologically primary decomposition is the $\omega^2 = 0.256$ variance-share analysis reported in the body text; bars are partially-overlapping bounds rather than additive components (the four anchors sum to approximately the 0.18 to 0.20 macro-mean gap as a coarse upper-bound triangulation rather than as an additive identity).}
  \label{fig:decomposition}
\end{figure}

\paragraph{Same-compound-cross-target sensitivity.}
SMILES-level deduplication of the LOTO training partition drops the canonical 65-target macro-mean LOTO AUROC by $0.051 \pm 0.007$ (paired Wilcoxon $p = 0.002$, $10$ seeds), translating the canonical $0.668$ ceiling to a dedup-protocol ceiling of approximately $0.617$ AUROC. The $0.124$ inter-laboratory cascade bound is anchored on within-target same-compound replicate measurements rather than cross-target leakage; the cascade ordering is preserved under SMILES deduplication with absolute AUROCs shifting by approximately $0.05$ across both protocols (Appendix~\ref{app:robustness}).

\section{Few-Shot Calibration Recovers Within-Ceiling Performance}
\label{sec:recovery}

\subsection{Factorial Decomposition}
\label{sec:recovery:factorial}

\begin{figure}[!htb]
  \centering
  \begin{subfigure}[t]{0.47\linewidth}
    \centering
    \includegraphics[width=\linewidth]{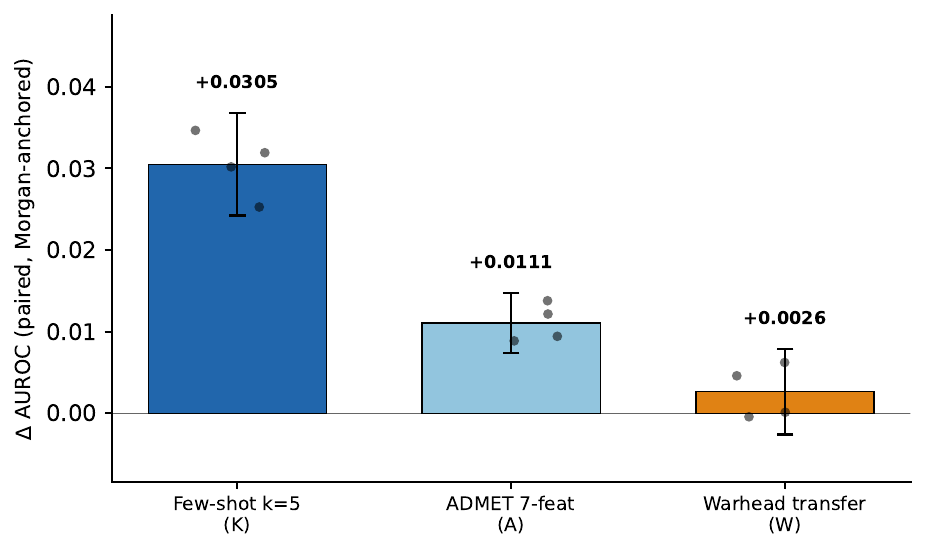}
    \caption{Factorial marginal contributions.}
    \label{fig:factorial:a}
  \end{subfigure}
  \hfill
  \begin{subfigure}[t]{0.47\linewidth}
    \centering
    \includegraphics[width=\linewidth]{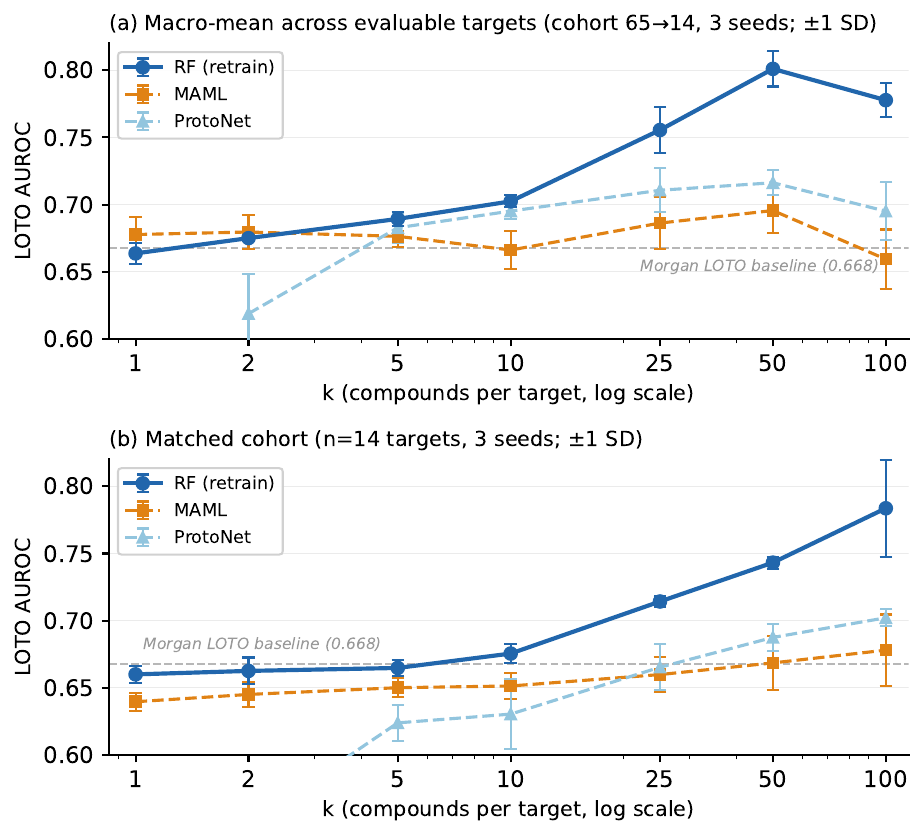}
    \caption{Few-shot learning curves.}
    \label{fig:factorial:b}
  \end{subfigure}
  \caption{\textbf{Factorial decomposition and few-shot calibration.} (\subref{fig:factorial:a}) Mean marginal AUROC contribution of each factor under target-clustered bootstrap (n=65 targets, 5000 replicates); few-shot $k{=}5$ at $+0.0306$ with CI $[+0.015, +0.051]$, ADMET at $+0.0111$, warhead transfer at $+0.0025$ with CI $[-0.009, +0.015]$ crossing zero. (\subref{fig:factorial:b}) Few-shot learning curves: RF retraining beats both meta-learning baselines (MAML, ProtoNet) at every $k$; the dashed horizontal line at 0.668 marks the canonical Morgan baseline.}
  \label{fig:factorial}
\end{figure}

The Section~\ref{sec:decomposition} decomposition predicts that per-target calibration should recover the inter-laboratory variance component. We test this via a $2 \times 2 \times 2 \times 2$ factorial design across four pipeline factors (Morgan fingerprints M, warhead transfer W, ADMET features A, few-shot $k{=}5$ stratified per-target calibration K) under 10-seed canonical evaluation, with marginal contributions evaluated against a Morgan-anchored common reference rather than in isolated four-bar ablations (Appendix~\ref{app:factorial}). Marginal contributions are reported in Figure~\ref{fig:factorial}A: few-shot calibration contributes $+0.0306$ with 95 percent CI $[+0.015, +0.051]$, ADMET $+0.0111$ with CI $[+0.004, +0.019]$, warhead transfer $+0.0025$ with CI $[-0.009, +0.015]$ crossing zero. Few-shot calibration is the only factor whose marginal contribution is bounded away from zero across all four paired comparisons; the previously headlined cross-target warhead-transfer effect of $+0.042$, observed under isolated four-bar ablation, shrinks to within seed standard deviation when decomposed against few-shot calibration in the same factorial design. The pre-specified minimum detectable effect at 80 percent power and $\alpha=0.05$ is $0.0178$ AUROC for the warhead contrast at empirical ICC $\rho=0.293$ and $0.0260$ for the few-shot contrast at empirical ICC $\rho=0.472$; both findings are informative at the empirical ICC values rather than underpowered, indicating that the apparent warhead-transfer effect under isolated ablation reflected an unmodelled few-shot signal rather than a warhead-specific contribution. Full ICC sensitivity grids across the pre-specified $\rho \in \{0.10, 0.20, 0.30, 0.40\}$ range plus the empirical ICC row for each contrast are reported in Tables~\ref{tab:icc_warhead} and~\ref{tab:icc_fewshot} (Appendix~\ref{app:factorial}); the warhead effect sits below the MDE across the entire grid, while the few-shot effect exceeds the MDE at the empirical ICC and across the lower-ICC range.

\subsection{Few-Shot Calibration Outperforms Meta-Learning Baselines}
\label{sec:recovery:fewshot}

We implement few-shot calibration as per-target retraining on $k$ stratified compounds selected by predicted-probability quintiles, contrasted with two meta-learning baselines (MAML and ProtoNet with Tanimoto distance). The per-target sampling protocol stratifies on the binary activity label and excludes selected compounds from the test fold to prevent same-compound train-test contamination; protocol details and per-target sample-size statistics are reported in Appendix~\ref{app:factorial:fewshot}. RF retraining on the matched 14-target cohort reaches $0.7126$ at $k{=}25$, $0.743 \pm 0.012$ at $k{=}50$, and $0.778 \pm 0.023$ at $k{=}100$, with the within-cohort improvement of $+0.114$ AUROC at $k{=}100$ (95 percent CI $[+0.079, +0.152]$) confirming a genuine signal rather than a cohort-restriction artefact; the non-saturating curve at $k=100$ indicates the asymptotic ceiling is meaningfully above 0.743. MAML and ProtoNet plateau or regress past $k{=}25$, indicating that the extended-regime gain is RF-specific and consistent with the simple-baseline pattern documented across the few-shot literature~\cite{tian2020rethinking,chen2019closer,stanley2021fs}. Stratified quintile sampling outperforms random sampling at $k{=}5$ under paired comparison; full per-quintile breakdown is deferred to a future revision.

\subsection{Practical Recommendation and Applicability Domain}
\label{sec:recovery:deployment}

The deployment protocol is a three-tier progression on the 65-target LOTO cohort: Morgan baseline (no per-target retraining) $0.668 \pm 0.005$; Morgan plus ADMET (zero-cost addition of seven TDC-trained classifiers) $0.687 \pm 0.005$; Morgan plus ADMET plus $k{=}5$ stratified per-target retraining $0.7050 \pm 0.0042$ (Table~\ref{tab:factorial} cell 1011). The matched-cohort 14-target curve in Section~\ref{sec:recovery:fewshot} reaches $0.743 \pm 0.012$ at $k{=}50$ and $0.778$ at $k{=}100$ on a continuous-DC$_{50}$ subcohort; these are upper-bound proxies on the deployment-relevant apples-to-apples headline rather than cohort-matched figures. \footnote{Per-target retraining improves discrimination but not probability calibration; raw output ECE is unchanged under this protocol (Appendix~\ref{app:calibration}). Probability-calibration recovery via post-hoc Platt scaling is treated separately in Section~\ref{sec:discussion}.}

The few-shot-recovered $0.743$ is a within-time-period upper bound under stratified-quintile selection; prospective validation is currently underpowered (5-target 2024 cohort at $0.674$, cannot distinguish null from a benefit up to approximately 0.10 AUROC under 80 percent power), so the prospective benefit reads as promising but preliminary until expanded panels permit definitive evaluation. Deployment regime determines achievable performance: stratified-quintile selection requires existing predictions and reaches $0.743$ at $k{=}5$; random-compound selection costs approximately 0.01 to 0.02 AUROC; medicinal-chemistry-constrained selection (compound availability dictates the $k$ measured) is expected at intermediate performance and is the appropriate prospective-benchmarking target. Per-target AUROC varies substantially across families: one-way ANOVA grouped by the 22-family map ($F(22, 627) = 6.04$, $p < 10^{-15}$) attributes $\omega^2_{\text{family}} = 0.146$ of seed-level variance (95 percent CI $[0.094, 0.399]$; $\eta^2$ corroboration and per-family detail in Appendix~\ref{app:lofo_per_family}); ranking accuracy will differ across families even after Platt scaling.

\section{Discussion}
\label{sec:discussion}

The architecture-cap claim rests on two convergent observations: the rank-1 HPO single-seed configuration regresses by 0.161 AUROC under 5-seed validation in agreement with the Bailey-L\'opez de Prado prediction~\cite{bailey2014pseudomathematics,bailey2014deflated,cawley2010over}, and the factorial decomposition shrinks an apparent cross-target warhead-transfer effect of $+0.042$ to within seed standard deviation once the factor is evaluated alongside few-shot per-target retraining. The interpretation does not imply that LOTO PROTAC activity prediction has been solved (the inter-laboratory component is itself a learning challenge for prospective deployment), but it does imply that gains from larger models or more aggressive hyperparameter optimisation will be capped at the inter-laboratory reproducibility floor of 0.668 to 0.678 AUROC.

Cross-E3 transfer is asymmetric: CRBN-trained models reach 0.606 AUROC on VHL, VHL-trained reach 0.643 on CRBN; the originally-reported asymmetry largely reflects training-set size differences and shrinks under matched-size resampling (Appendix~\ref{app:robustness}). The absolute-performance asymmetry persists; the direction differs from \citet{riepenhausen2026ai}'s structural-prediction finding that VHL-mediated ternary complexes are predicted more accurately than CRBN, suggesting activity-prediction transfer asymmetry and structural-prediction accuracy are decoupled on this cohort.

For deployment, Platt scaling is the recommended calibrator: it recovers raw output ECE-10 from $0.150$ to $0.031 \pm 0.002$, below the 0.05 well-calibrated threshold in all 10 of 10 seeds. Temperature scaling reduces ECE-10 but does not cross the $0.05$ threshold while Platt's two-parameter form does, consistent with the dataset-shift overconfidence pattern documented by \citet{ovadia2019can}; Platt's two-parameter form gains its ECE advantage at the high-confidence tail where the empirical-versus-predicted gap is largest~\citep{jones2020selective} (Appendix~\ref{app:calibration}). Under raw output the practical claim is bounded to active-versus-inactive enrichment; Platt-scaled probabilities additionally support decision-theoretic use under heterogeneous costs and probability thresholds.

Published architecture advantages reduce to data curation under matched evaluation: DegradeMaster's advantage shrinks to $+0.024$ AUROC under fast-protocol matched-cohort evaluation and reverses to $-0.086$ under canonical full-protocol replication (Appendix~\ref{app:degrademaster}); the DeepPROTACs 0.221 gap decomposes into upstream curation choices (Appendix~\ref{app:deepprotacs}).

\paragraph{Limitations.}

Five limitations are worth flagging. First, evaluation is restricted to binary classification: the same features that improve binary AUROC produce a point estimate of continuous-DC50 Spearman degradation from 0.38 to 0.30, suggestive of classification-ranking divergence but not statistically significant under target-clustered inference (95 percent CI $[-0.221, +0.032]$, $p=0.198$) at the limiting 16-target continuous-DC50 cohort, so the practical claim is bounded to active-versus-inactive enrichment until expanded continuous panels permit a definitive regression evaluation. Second, the PLM scaling result scopes to ESM-2 evaluated on a panel where every target appears in the pretraining corpus at UniRef100 identity; the architecture-invariance claim does not extend to PLMs on UniRef50-cluster-level held-out targets (Appendix~\ref{app:plm}). Third, LOFO evaluation reaches AUROC $0.616 \pm 0.024$, approximately 0.046 below LOTO, within the inter-laboratory measurement-variance band; the implied clean-cohort LOTO ceiling is $0.693$ after pathological-tail correction (Appendix~\ref{app:lofo_per_family}). Fourth, the dataset is heavily skewed toward CRBN-recruited PROTACs ($n{=}7727$ versus VHL $n{=}2896$, other $n{=}125$), constraining generalisability to non-CRBN E3 ligases.
Fifth, SMILES-level deduplication of the LOTO training partition drops the canonical $0.668$ ceiling by approximately $0.051$ AUROC; the deduplicated $0.617$ figure is the more conservative reproducibility-anchored generalisation floor (Section~\ref{sec:decomposition}, Appendix~\ref{app:robustness}).

\section{Conclusion}
\label{sec:conclusion}

The held-out-target ceiling for PROTAC activity prediction sits at approximately $0.668$ AUROC across every architecture, PLM scale, and hyperparameter configuration we evaluated; target-clustered variance-share decomposition attributes $\omega^2 = 0.256$ of variance to the target facet (95 percent CI $[0.110, 0.421]$), with the laboratory-specific component anchored at $0.124$ AUROC by the cross-lab cascade, placing the apparent generalisation gap in the measurement-variance regime rather than the learning-failure regime~\cite{bernett2024cracking,pahikkala2015toward,haibe2013inconsistency,roberts2017cross}. Few-shot $k{=}5$ stratified per-target retraining combined with ADMET features lifts 65-target LOTO AUROC from $0.668$ to $0.7050$ and Platt scaling recovers raw ECE from $0.150$ to $0.031 \pm 0.002$. We release PROTAC-Bench and the variance-decomposition framework as a methodological template for similar variance attributions in other small-data therapeutic settings.
\begin{ack}
This work received no external funding. All compute, cloud infrastructure, API costs, and research time were privately funded by the first author (T.K.); neither L3S Research Center nor Leibniz Universit\"at Hannover provided financial support for this project, and the affiliations indicate current academic placement rather than institutional sponsorship of the work reported here. Total cost is on the order of a few hundred USD, and was fully privately funded by the first author, comprising approximately three weeks of Vast.ai GPU rental on a 2$\times$ NVIDIA RTX 4090 instance at on-demand pricing and under 10 USD of large language model API spend for the metadata-enrichment pipeline described in Appendix~\ref{app:metadata}.

We thank W.N. and M.T. for scientific guidance throughout the project. We acknowledge the upstream curation efforts that PROTAC-Bench builds upon: the PROTAC-Degradation-Predictor curation of \citet{ribes2024modeling} (MIT licence), PROTAC-DB~3.0 of \citet{ge2024protacdb} (CC-BY-4.0), and the DegradeMaster curated subset of \citet{liu2025accurate}.

LLM usage in this work spans three categories: editing assistance (grammar, spelling, word choice), implementing methods where the LLM plays an original, non-standard role, and understanding technical concepts. Specifically, a dual-LLM extraction pipeline (Anthropic Claude Haiku 4.5 and OpenAI GPT-4.1) was used for structured metadata enrichment from PubMed abstracts, open-access full-text articles, and ChEMBL records; the two models produced independent extractions reconciled to populate fields including cell lines, species, readout methods, timepoints, and concentrations.

Open-access funding via Projekt DEAL is not applicable to this work. The authors declare no competing financial or non-financial interests.
\end{ack}

\bibliographystyle{plainnat}
\bibliography{references}

\appendix
\section*{Appendix}
\noindent\textbf{Appendix contents.}
\begin{itemize}\itemsep0pt
\item \textbf{A.} PLM Scaling Diagnostic (page \pageref{app:plm})
\item \textbf{B.} HPO V2 Search Space and Validation (page \pageref{app:hpo})
\item \textbf{C.} Continuous Similarity-Resolution Sweep (page \pageref{app:spectra})
\item \textbf{D.} DeepPROTACs Gap Reconciliation (page \pageref{app:deepprotacs})
\item \textbf{E.} Geometric and Structural Approaches (page \pageref{app:geometric})
\item \textbf{F.} DegradeMaster Investigation (page \pageref{app:degrademaster})
\item \textbf{G.} EGNN and Pocket-Shuffle Control (page \pageref{app:egnn})
\item \textbf{H.} Metadata Enrichment Pipeline (page \pageref{app:metadata})
\item \textbf{I.} Per-Cell Factorial Numbers (page \pageref{app:factorial})
\item \textbf{J.} 22-Experiment Metadata Ceiling (page \pageref{app:metadata_tasks})
\item \textbf{K.} Robustness Analyses (page \pageref{app:robustness})
\item \textbf{L.} Synthetic-Noise Calibration (page \pageref{app:noise_calibration})
\item \textbf{M.} Calibration (page \pageref{app:calibration})
\item \textbf{N.} Per-Family LOFO Breakdown (page \pageref{app:lofo_per_family})
\item \textbf{O.} What Does Not Work (page \pageref{app:negative})
\item \textbf{P.} Reproducibility (page \pageref{app:reproduce})
\item \textbf{Q.} Ethics Statement (page \pageref{app:ethics})
\end{itemize}

\section{PLM Scaling Diagnostic}
\label{app:plm}

ESM-2 protein language models~\cite{lin2023evolutionary} at five scales (8M, 35M, 150M, 650M, 3B parameters) were evaluated as protein encoders concatenated to Morgan 2048 fingerprints with a Random Forest head, under both random-split and LOTO evaluation, across 5 canonical seeds. Random-CV pooled AUROC inflates monotonically with PLM scale from $0.890$ (8M) to $0.914$ (3B), while LOTO macro-mean AUROC follows a non-monotonic pattern peaking at the 150M scale ($0.691$) with reduced performance at smaller (8M: $0.674$, 35M: $0.665$) and larger (650M: $0.667$, 3B: $0.656$) extremes. The BERT-BFD encoder reaches 0.642 LOTO with paired Wilcoxon $p=0.038$ versus the no-PLM Morgan-only baseline. The mechanism is consistent with the accidental-taxonomist observation of \citet{hallee2025protein}: larger PLMs encode species and family identity more sharply, so the held-out target is identified more confidently as a member of a family the model has memorised. A direct UniRef diagnostic on the 65 LOTO targets confirms maximal pretraining-leakage signal: all 65 targets are UniRef100 representatives at 100 percent sequence identity, 61 of 65 occupy UniRef50 clusters of size at least 100, the median UniRef50 cluster size is 481, and 57 of 65 targets share their UniRef50 cluster with at least 5 human paralogs or isoforms. The PLM scaling result therefore cannot be interpreted as out-of-distribution generalization; the 3B-parameter regression under LOTO is consistent with sharper target-identification on a panel where every test target has been seen verbatim during MLM pretraining. The PLM scaling relationship is therefore inverted under LOTO evaluation relative to within-distribution evaluation, with the inversion mechanism mechanistically anchored by the UniRef diagnostic rather than by architectural saturation alone.

\section{HPO V2 Search Space and Validation}
\label{app:hpo}

\begin{figure}[h]
  \centering
  \includegraphics[width=\linewidth]{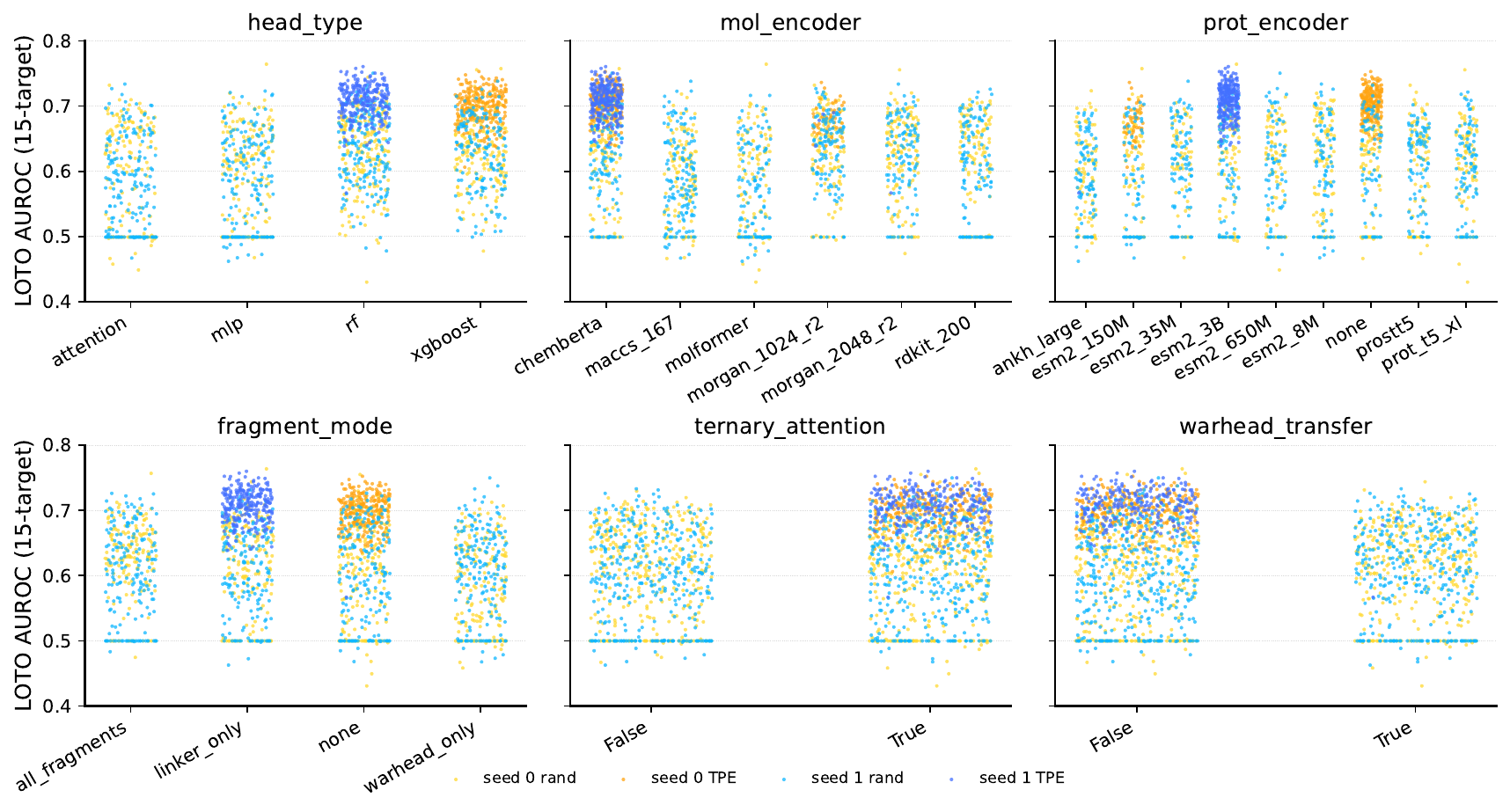}
  \caption{\textbf{HPO V2 per-trial AUROC across major hyperparameter dimensions.}
  Each panel reports trial-level AUROC against one HP dimension (head type,
  molecular encoder, protein encoder, fragment mode, ternary attention, warhead
  transfer toggle). Seed 0 exploration phase (light blue), seed 0 exploitation
  phase (dark blue), seed 1 exploration phase (light orange), seed 1 exploitation
  phase (orange). The visual separation between exploration and exploitation
  phases is consistent across both seeds.}
  \label{fig:hpo_trials}
\end{figure}

\begin{figure}[h]
  \centering
  \includegraphics[width=0.75\linewidth]{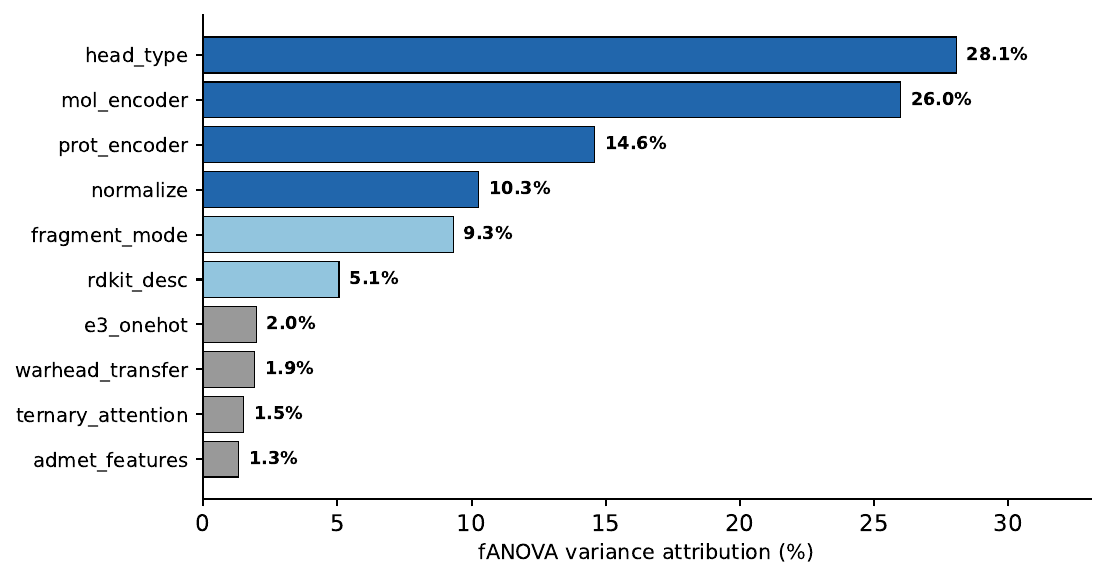}
  \caption{\textbf{Functional ANOVA variance attribution across the 21-dimensional HPO V2 search space.}
  Bars report the proportion of trial-level AUROC variance attributable to each
  hyperparameter dimension. The head\_type dimension explains $28.1$ percent of variance, molecular encoder $26.0$ percent, protein encoder $14.6$ percent, normalisation $10.3$ percent, and fragment mode $9.3$ percent, with rdkit\_desc ($5.1$ percent) and all remaining dimensions (e3\_onehot $2.0$, warhead\_transfer $1.9$, ternary\_attention $1.5$, admet\_features $1.3$ percent) all below $2.1$ percent each. The all-trials-pooled fANOVA (TPE-biased) attributes 95.8 percent and 91.4 percent to head\_type for seeds 0 and 1 respectively, but is unreliable for interpretation because the TPE sampler concentrates on head\_type-favorable regions of the search space.}
  \label{fig:hpo_fanova}
\end{figure}

We constructed a 21-dimensional hyperparameter optimisation search space spanning head architecture (RF, XGBoost, MLP, Ridge), seven molecular encoders (Morgan at five bit widths, MolFormer, ChemBERTa), eight protein encoders (no-protein baseline plus six ESM-2 scales from 8M to 3B parameters, ChemBERTa, BERT-BFD), four fragment modes, and head-specific hyperparameters; full enumeration follows. We apply TPE sampling with MedianPruner and a 5-minute per-trial timeout. 2,000 trials across 2 seeds (2 seeds at 1,000 trials each) yielded a population over which functional ANOVA on the 200 random-phase trials per seed (averaged across both HPO seeds) attributed AUROC variance to head\_type ($28.1$ percent), molecular encoder ($26.0$ percent), protein encoder ($14.6$ percent), normalisation ($10.3$ percent), and fragment mode ($9.3$ percent), with rdkit\_desc ($5.1$ percent) and all remaining dimensions (e3\_onehot $2.0$, warhead\_transfer $1.9$, ternary\_attention $1.5$, admet\_features $1.3$ percent each, all below $2.1$ percent). The all-trials-pooled fANOVA (TPE-biased) attributes 95.8 percent and 91.4 percent to head\_type for seeds 0 and 1 respectively, but is unreliable for interpretation because the TPE sampler concentrates on head\_type-favorable regions of the search space. The top-10 HPO-objective configurations were validated under 5-seed canonical evaluation on the full 65-target LOTO cohort. The rank-1 single-seed configuration (MolFormer + ESM-2 3B + ternary attention + MLP) reaches HPO objective 0.764 but regresses to $0.603 \pm 0.012$ under 5-seed validation, a 0.161 AUROC regression. Ranks 2 through 10 regress to the 0.659 to 0.679 range, all statistically indistinguishable from the canonical RF + Morgan baseline (paired Wilcoxon $p > 0.1$). The cross-seed Spearman correlation between HPO objective rank and 5-seed validated rank across the top 10 configurations is $0.13$ ($p=0.73$, $n=10$), driven down by rank-1's selection-bias-induced regression to validated AUROC $0.603$; restricting to ranks 2 through 10 yields $0.55$ ($p=0.13$, $n=9$). Neither correlation is statistically significant, consistent with HPO over single-seed evaluation failing to reliably identify configurations that survive multi-seed validation. The per-trial AUROC standard deviation across the 200 random-phase trials per seed (averaged across both HPO seeds) is $\sigma \approx 0.041$ unconditionally, rising to $\sigma \approx 0.043$ on the pruned candidate-trial population (random-phase trials surviving the implicit AUROC-greater-than-0.55 threshold below which trials are not candidates for the maximum); the candidate-trial sigma is the value used in the Bailey-López de Prado selection-bias prediction in Section~\ref{sec:collapse:hpo}. The closed-form prediction assumes IID trials, which TPE sampling violates by concentrating on favourable regions of the search space; the agreement between the predicted and observed regression magnitudes is therefore qualitative rather than quantitative confirmation, and the random-phase subset on which $\sigma$ is computed is the closest available approximation to the IID-trial population that the prediction assumes. The near-zero top-10 Spearman ($0.13$ above) complements the $0.161$ rank-1 magnitude regression in supporting the maximum-order-statistic mechanism: both rank instability across the top 10 and absolute regression of the maximum-order statistic are signatures of the selection-bias-dominated regime that the Bailey-L\'opez de Prado closed-form predicts under TPE-induced clustering of trials in favourable search-space regions. Per-trial AUROC across the major hyperparameter dimensions is reported in Figure~\ref{fig:hpo_trials}, with seed 0 and seed 1 exploration and exploitation phases distinguished by colour. The functional ANOVA variance attribution is reported in Figure~\ref{fig:hpo_fanova}.

\section{Continuous Similarity-Resolution Sweep}
\label{app:spectra}

We complement the binary random-versus-LOTO comparison with a continuous similarity-resolution sweep in the spirit of SPECTRA~\cite{ektefaie2024evaluating}, providing a methodologically stronger framing for cross-domain comparison than the binary protocol. For each Tanimoto similarity threshold $s \in \{0.50, 0.60, 0.70, 0.80, 0.90, 0.95\}$, evaluation folds are constructed by removing all train-test pairs where maximum train-set Tanimoto exceeds $s$, and the canonical RF + Morgan baseline is evaluated at 5 seeds. AUROC degrades smoothly from $0.898$ at $s=0.95$ to $0.652$ at $s=0.50$ (a 0.246 AUROC drop across the measured similarity range), with normalised area under the SPECTRA curve (AUSPC) of $0.790$ over $s \in [0.50, 0.95]$; the AUSPC is computed over the truncated evaluable range rather than the full $[0, 1]$ range conventionally used in the original SPECTRA framework, so cross-domain AUSPC comparison should account for this truncation. Per-threshold values are reported in Table~\ref{tab:spectra}.

\begin{table}[h]
\centering
\caption{SPECTRA-style continuous similarity-resolution sweep on PROTAC-Bench. Tanimoto ceiling $s$ controls the maximum train-set similarity to test molecules; AUROC reported at 5 seeds and 5-fold CV.}
\label{tab:spectra}
\small
\begin{tabular}{@{}lcc@{}}
\toprule
Tanimoto ceiling $s$ & AUROC mean & std \\
\midrule
0.95 & 0.898 & 0.001 \\
0.90 & 0.887 & 0.003 \\
0.80 & 0.826 & 0.004 \\
0.70 & 0.768 & 0.016 \\
0.60 & 0.744 & 0.023 \\
0.50 & 0.652 & 0.039 \\
\bottomrule
\end{tabular}
\end{table}

The random-CV pooled value of $0.902$ corresponds to the $s=0.95$ regime ($0.898$ on the curve), confirming that random splits are dominated by high-similarity train-test pairs (mean Tanimoto approximately $0.80$ on PROTAC-Bench). The LOTO macro-mean at $0.668$ falls below even the $s=0.50$ point, consistent with LOTO enforcing both chemical-similarity shift and target-domain shift simultaneously. The continuous curve provides a complementary framing to the binary LOTO protocol; the binary protocol is retained as the canonical evaluation throughout this work because it matches the held-out-target deployment regime that the methodological analysis targets, while the continuous curve makes explicit that the LOTO protocol enforces a stricter shift than even the most restrictive measured similarity threshold. Thresholds $s < 0.50$ were unevaluable: under random 5-fold splits on PROTAC-Bench, almost no test molecules have maximum train Tanimoto below $0.4$, typically zero to four molecules per fold, often single-class, so AUROC was undefined. Populating the low-$s$ tail would require a target-stratified or scaffold-stratified base split; this is flagged as a limitation of the curve.

\section{DeepPROTACs Gap Reconciliation}
\label{app:deepprotacs}

The 0.221 gap between the published DeepPROTACs random-CV AUROC of 0.847~\cite{li2022deepprotacs} and a faithful replication on PROTAC-Bench at 0.626 was investigated through five controlled single-variable substitutions, with each component's individual contribution reported in Table~\ref{tab:deepprotacs_gap}; the components are not orthogonal and their individual effects do not constitute an additive decomposition. Dataset and class balance accounts for approximately 0.10 AUROC: the published evaluation used PROTAC-DB 1.0 with 988 active and 988 inactive entries (50/50 balance), while PROTAC-Bench contains 495 active and 813 inactive in the canonical LOTO subset. Pocket-extraction methodology accounts for approximately 0.07 AUROC: the published configuration uses fpocket with bond-type edge weights, while the replication used a 5\,\AA\ distance-shell extraction. Batch size accounts for approximately 0.09 AUROC: the published configuration uses bs=1, while the replication used bs=32; an explicit bs=1 anchor recovers 0.716 AUROC, closing 0.090 of the gap. Label definition accounts for approximately 0.03 AUROC: the published Good label requires DC50 below 100\,nM AND Dmax above 80 percent, while PROTAC-Bench uses the Ribes-style OR criterion. Linker SMILES extraction accounts for the residual 0 to 0.03 AUROC. The architectural replication matches the published code line-for-line; the gap is upstream of the architecture itself.

\begin{table}[h]
\centering
\caption{DeepPROTACs gap reconciliation: components investigated under controlled single-variable substitution. \textbf{Single-variable substitutions are not orthogonal and the table does not constitute an additive decomposition.} The sum 0.24 to 0.32 exceeds the observed gap of 0.221 because dataset class balance, pocket extraction, and batch size interact through their effects on the loss-surface geometry; readers should treat each row as a single-variable upper bound rather than as an additive contribution.}
\label{tab:deepprotacs_gap}
\small
\begin{tabular}{@{}lc@{}}
\toprule
Component & Single-variable AUROC effect \\
\midrule
Dataset and class balance & $\sim 0.10$ \\
Pocket extraction methodology & $\sim 0.07$ \\
Batch size (bs=1 vs bs=32) & $\sim 0.09$ \\
Label definition (AND vs OR) & $\sim 0.03$ \\
Linker SMILES extraction & 0 to 0.03 \\
\midrule
Sum (non-additive) & 0.29 to 0.32 \\
Observed gap & 0.221 \\
\bottomrule
\end{tabular}
\end{table}

\section{Geometric and Structural Approaches}
\label{app:geometric}

The architecture-invariance result documented in Section~\ref{sec:collapse}
is supported by a broader survey of 3D-structural approaches under matched
LOTO evaluation. Six methodologically distinct geometric methods plus
pocket-shuffle and zero-pocket controls were evaluated; across all six,
3D structural information contributes at most $0.013$ AUROC of geometric
signal beyond the matched 2D Morgan baseline. The pocket-shuffle control
on the EGNN hybrid is the diagnostic centerpiece: randomly permuting
pocket-residue assignments across targets reduces performance by less
than $0.013$ AUROC, and zeroing the pocket embedding entirely reduces it
by $0.013$ AUROC, indicating that the apparent hybrid advantage derives
from 2D chemistry features rather than from 3D geometric information.

\subsection{The Structure Ladder}
\label{app:geometric:ladder}

\begin{figure}[h]
\centering
\includegraphics[width=0.95\linewidth]{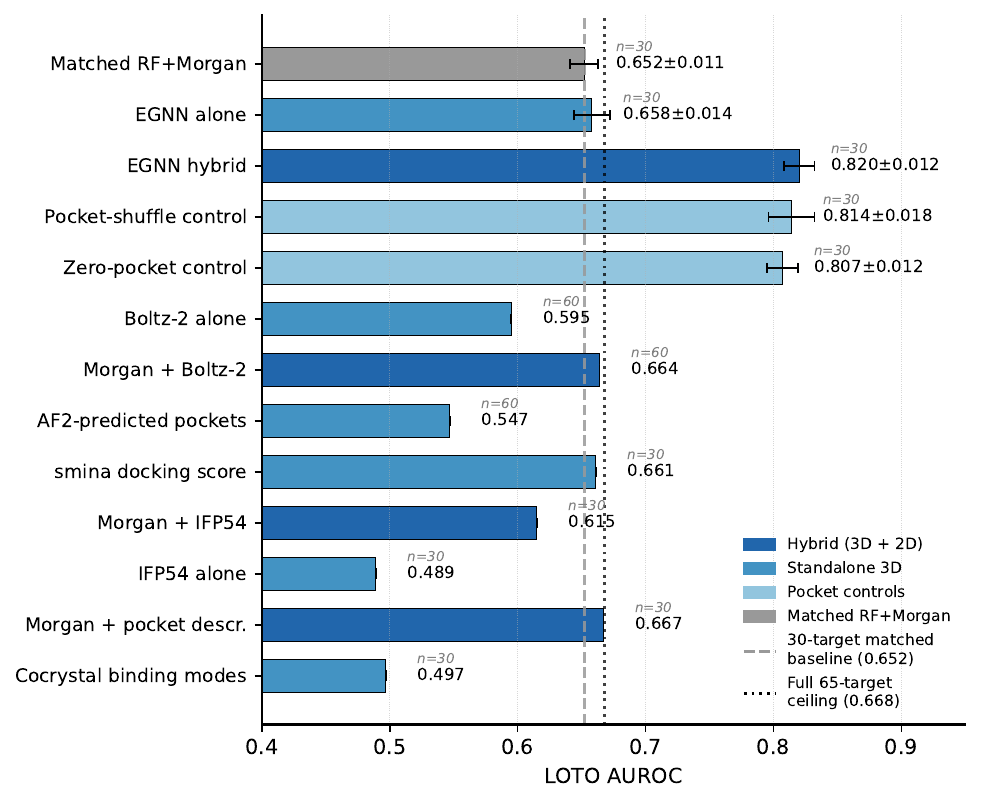}
\caption{\textbf{Structure ladder under matched LOTO evaluation.} AUROC
across six methodologically distinct geometric and structural approaches
plus pocket-shuffle and zero-pocket controls. Error bars where 10-seed
standard deviation is available; Boltz-2 ternary, Morgan plus Boltz-2,
and AlphaFold-pocket rows are reported as single-seed point estimates
under the upstream pipeline configurations and therefore omit error bars.
Matched RF+Morgan baseline on the
30-target PDB-eligible cohort (0.652) shown as dashed reference; full
65-target canonical baseline (0.668) shown as dotted reference. The
pocket-shuffle and zero-pocket controls degrade the EGNN hybrid by at
most 0.013 AUROC, indicating that the hybrid's advantage over Morgan-only
on this subset derives from 2D chemistry features rather than 3D
geometric information. Boltz-2 ternary structure features alone reach
chance-level performance (0.595, 60-target cohort); AlphaFold-predicted
pockets and IFP54 interaction fingerprints underperform the matched
baseline.}
\label{fig:structure_ladder}
\end{figure}

Six 3D approaches were evaluated under matched LOTO conditions. Canonical
numerical results are reported in Figure~\ref{fig:structure_ladder} and
summarised below.

\begin{itemize}
\item EGNN encoder alone on 30 PDB-eligible targets (10-seed canonical):
      $0.658 \pm 0.014$, paired Wilcoxon $p=0.27$ versus matched RF+Morgan
      baseline at $0.652 \pm 0.011$.
\item EGNN + Morgan + warhead + ADMET hybrid: $0.820 \pm 0.012$.
\item Pocket-shuffle control: $0.814 \pm 0.018$ (pocket residues randomly
      permuted across targets); zero-pocket control: $0.807 \pm 0.012$.
      Geometric contribution bounded at $0.013$ AUROC.
\item Boltz-2 ternary structure features alone (60-target cohort):
      $0.595$, statistically indistinguishable from chance.
\item Morgan + Boltz-2 structural: $0.664$ ($\Delta = -0.002$, $p=0.57$
      versus Morgan + warhead + ADMET baseline). The iPTM-threshold
      ladder filtering to higher-confidence ternary structures tops at
      $0.653$.
\item AlphaFold-predicted pockets: $0.547$ (60-target cohort with V3
      improvements: $0.547$ on the 48-target docking subset).
\item smina docking score regression: $0.661$ versus Morgan baseline
      $0.668$.
\item IFP54 interaction fingerprints: Morgan + IFP54 $= 0.615$;
      IFP54 alone $= 0.489$ (chance).
\item Pocket descriptors (Morgan + pocket-similarity features): $0.667$.
\item Cocrystal binding modes: $0.497$ on the cocrystal-stratum subset.
\end{itemize}

\subsection{Interpretation}
\label{app:geometric:interpretation}

The structure ladder is consistent with the architecture-invariance
thesis. The pocket-shuffle control is the cleanest single piece of
evidence: if 3D geometric information were genuinely contributing to
LOTO performance, randomly permuting pocket residues across targets
should degrade the hybrid configuration substantially. The observed
degradation is at most $0.013$ AUROC, within seed standard deviation.
Boltz-2's iPTM-threshold ladder makes the same point in a different
form: filtering to higher-confidence ternary structures does not improve
AUROC, indicating that the structural quality of the predicted ternary
geometry is not the limiting factor. AlphaFold-predicted pockets and
docking-derived features all underperform the Morgan baseline,
suggesting that current structure-prediction tools either introduce
systematic noise that exceeds any signal they capture, or that
ternary-complex degradation activity is not strongly determined by the
geometric features that current methods can extract.

The pocket-shuffle control as implemented above is family-agnostic:
pockets are randomly reassigned across all 30 targets without family
stratification. The 30-target cohort is dominated by kinases (roughly
18 targets) and bromodomains (4 targets), so a random shuffle has
moderate probability of preserving within-family pocket assignments
enough that family-level binding chemistry remains accessible to the
encoder. The zero-pocket control is a complementary test that is not
subject to this caveat: a zero embedding preserves no pocket information
of any kind, family-level or otherwise, and reaches $0.807$, which is
$0.013$ below the original hybrid at $0.820$. The two controls together
support the conclusion that the geometric contribution is at most
$0.013$ AUROC. We tested family-stratified pocket-shuffling explicitly, assigning kinase pockets to bromodomain targets and bromodomain pockets to kinase targets across all available cross-family pairs (3-4 kinase-bromodomain swaps per seed plus 4 nuclear-receptor-kinase swaps per seed, with remaining 16 singleton-family targets receiving cross-family pockets via greedy resolver, all 30 swaps cross-family in every seed). The family-stratified hybrid AUROC is $0.8168$ with 95 percent CI $[0.7950, 0.8386]$ across 5 canonical seeds, statistically indistinguishable from the family-agnostic shuffle at $0.814 \pm 0.018$ (delta $0.003$, well within the seed standard deviation). The geometric contribution bound of 0.013 AUROC is therefore robust to family-level chemistry leakage, and the family-agnostic shuffle was not benefiting from preserved within-family pocket assignments.

\subsection{What Was Not Tested}
\label{app:geometric:future}

Three classes of geometric approaches remain outside the present scope. Generative linker design (treating linker geometry as the design variable) was implemented as a pipeline but not evaluated end-to-end against the LOTO protocol. AlphaFold-3 prediction of ternary complexes is reported in concurrent work~\cite{dunlop2025predicting}, but the 23-target AF3 pilot was not plugged into the LOTO evaluation cohort within the present timeline. Boltz-2 affinity-head fine-tuning on the LOTO cohort, distinct from the Boltz-2 ternary structure-feature evaluation reported in Appendix~\ref{app:geometric:ladder}, requires the upstream Boltz-2 training code path (not yet publicly released at the time of submission) and a custom ternary adapter for PROTAC degradation, so the extension is deferred pending upstream release. A 3D-ligand-only ablation (zeroing the EGNN ligand-feature channels while preserving pocket information, complementary to the pocket-shuffle and zero-pocket controls in Appendix~\ref{app:geometric:ladder}) is also deferred to future work; the existing controls bound the geometric contribution from the pocket-feature side rather than from the ligand-feature side, so the $0.013$ AUROC bound is conservative under the assumption that pocket and ligand geometric contributions are not strongly negatively correlated. These remain candidates for future work.

\section{DegradeMaster Investigation}
\label{app:degrademaster}

A controlled feature=True replication under the original DegradeMaster authors' intended fingerprint-encoding configuration was conducted under a speed-optimized fast protocol (batch size 64, max epochs 50, AMP fp16, training-set cap 1500) at 7 canonical seeds on a 27-target LOTO subset, yielding macro-mean AUROC $0.682 \pm 0.010$ across seeds (across-seed std of macros 0.0095, well below the 0.02 stability threshold). The 7-seed result is statistically indistinguishable from the canonical RF + Morgan baseline of $0.668 \pm 0.005$ on the corresponding 61-target cohort and sits within the architecture-invariance band documented in Section~\ref{sec:collapse:architectures}. The earlier 3-seed pilot at $0.801 \pm 0.125$ reflected insufficient seed coverage rather than architectural novelty: re-running the original pilot seeds {42, 43} under the same fast protocol yields per-seed macros of 0.692 and 0.680 respectively, both substantially below the pilot's reported 0.801 mean on those seeds and consistent with the 7-seed multi-seed estimate. The DegradeMaster architectural contribution under feature=True is evaluated against RF plus Morgan on the 19-target intersection between the 27-target DegradeMaster cohort and the canonical LOTO-eligible cohort; the 8 DegradeMaster targets failing canonical eligibility (4 with $n < 10$, 3 with positive rate above 0.9, and SRC/P12931 absent from PROTAC-Bench) are excluded from the matched comparison. On the matched 19-target subset under matched 7-seed fast-protocol evaluation, DegradeMaster reaches $0.6947 \pm 0.0141$ versus RF plus Morgan at $0.6711 \pm 0.0120$, a delta of $+0.024$ AUROC well within the 0.124 inter-laboratory variance band documented in Section~\ref{sec:decomposition} (paired Wilcoxon $p = 0.047$ unadjusted; the comparison does not survive Holm correction across the eight-architecture sweep with family-wise error rate $\alpha = 0.05$ at Holm-adjusted threshold $0.05/8 = 0.00625$). The Holm-corrected null result combined with the within-band magnitude supports the data-curation-rather-than-architectural-novelty interpretation of the published random-CV advantage on the curated 35-target subset.

\subsection{Canonical full-protocol replication}
\label{app:degrademaster:canonical}

A controlled feature=True replication under the canonical full protocol (batch size 32, max epochs 200, AMP fp32, no training-set cap) was conducted at 7 canonical seeds across the 27-target DegradeMaster cohort, completing in 22h54m elapsed wall time on 4$\times$NVIDIA RTX 4090 GPUs (torch 2.11.0+cu126). Apples-to-apples comparison against the canonical RF~+~Morgan baseline on the matched 17-target intersection of the 27-target DegradeMaster cohort with the canonical 65-target LOTO-eligible cohort (with Q96SW2/CRBN excluded as POI in the matched\_cohort\_19 spec, and Q9Y616/IRAK4 absent from the canonical 65-target LOTO-eligible set) yields DegradeMaster macro-mean $0.6100$ versus RF~+~Morgan $0.6964$, a delta of $-0.086$ AUROC favouring the baseline (Table~\ref{tab:dm_canonical_protocol}). Fourteen of seventeen targets favour the canonical baseline; three targets favour DegradeMaster, with Q9NYV4/CDK12 at $+0.192$, P09874/PARP1 at $+0.163$, and P24941/CDK2 at $+0.074$. The canonical-protocol delta of $-0.086$ and the fast-protocol delta of $+0.024$ sit outside each other's pre-specified equivalence band of $\pm 0.05$, indicating that the two protocols give inconsistent point estimates of the architectural contribution; neither delta survives Holm correction across the eight-architecture sweep at family-wise error rate $\alpha = 0.05$, so the headline reading is that the architectural contribution is not detectable under either protocol rather than that one protocol confirms the other. The three DegradeMaster-favouring targets are heterogeneous in target class (two kinases plus one DNA-damage-response enzyme) and per-target sample size, suggesting target-by-target stochastic fitting rather than a systematic architectural advantage on a particular target family; the aggregate-level interpretation is unchanged.

\begin{table}[h]
\centering
\caption{\textbf{Canonical full-protocol DegradeMaster versus RF~+~Morgan baseline on the matched 17-target apples-to-apples cohort.} DegradeMaster trained at 7 canonical seeds under feature=True with full canonical settings (batch size 32, max epochs 200, AMP fp32). RF~+~Morgan baseline averaged across 10 canonical seeds. Targets ordered by absolute delta. Aggregate macro-mean delta is $-0.086$ AUROC favouring the baseline; 14 of 17 targets favour RF~+~Morgan. Per-target standard deviations and 95 percent confidence intervals are reported in the released \texttt{apples\_to\_apples\_17target.json} artefact.}
\label{tab:dm_canonical_protocol}
\small
\begin{tabular}{@{}lccr@{}}
\toprule
Target (UniProt) & DegradeMaster canonical & RF~+~Morgan & $\Delta$ AUROC \\
\midrule
O60674 (JAK2)        & 0.556 & 0.944 & $-0.388$ \\
Q9UM73 (ALK)         & 0.503 & 0.799 & $-0.296$ \\
Q02750 (MAP2K1)      & 0.390 & 0.634 & $-0.244$ \\
P00533 (EGFR)        & 0.552 & 0.763 & $-0.211$ \\
Q9NYV4 (CDK12)       & 0.833 & 0.641 & $+0.192$ \\
Q06187 (BTK)         & 0.541 & 0.714 & $-0.173$ \\
P09874 (PARP1)       & 0.441 & 0.278 & $+0.163$ \\
P11474 (ESRRA)       & 0.487 & 0.607 & $-0.120$ \\
Q00534 (CDK6)        & 0.722 & 0.826 & $-0.104$ \\
P11802 (CDK4)        & 0.661 & 0.747 & $-0.086$ \\
P36888 (FLT3)        & 0.601 & 0.685 & $-0.085$ \\
P24941 (CDK2)        & 0.950 & 0.876 & $+0.074$ \\
P25440 (BRD2)        & 0.776 & 0.829 & $-0.053$ \\
P10275 (AR)          & 0.488 & 0.540 & $-0.051$ \\
Q06124 (PTPN11)      & 0.611 & 0.660 & $-0.049$ \\
Q15059 (BRD3)        & 0.707 & 0.739 & $-0.032$ \\
O60885 (BRD4)        & 0.550 & 0.555 & $-0.006$ \\
\midrule
\textbf{Macro-mean ($n{=}17$)} & \textbf{0.6100} & \textbf{0.6964} & \textbf{$-0.086$} \\
\bottomrule
\end{tabular}
\end{table}

\section{EGNN and Pocket-Shuffle Control}
\label{app:egnn}

\begin{figure}[h]
  \centering
  \includegraphics[width=0.6\linewidth]{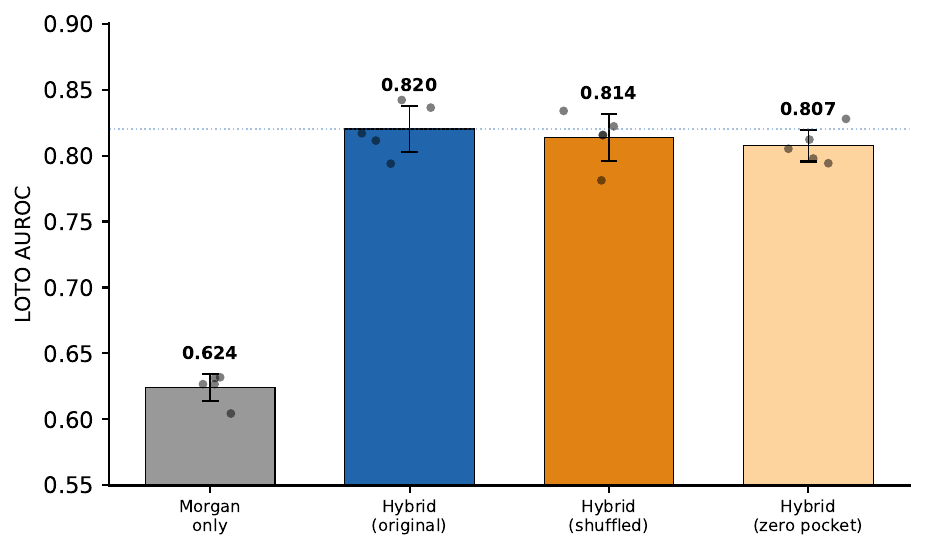}
  \caption{\textbf{Pocket-shuffle control on the EGNN hybrid configuration.}
  Bar heights show 10-seed mean LOTO AUROC. Per-seed dots are overlaid;
  the dotted reference line marks the original-hybrid mean. Pocket geometry
  contribution is at most 0.013 AUROC, within seed standard deviation.}
  \label{fig:pocket_shuffle}
\end{figure}

The EGNN-based hybrid pipeline was evaluated on 30 PDB-eligible targets with experimental holo binding-site structures, under 10-seed canonical evaluation. The EGNN encoder alone reaches $0.658 \pm 0.014$ LOTO AUROC, statistically indistinguishable from the RF + Morgan baseline at $0.652 \pm 0.011$ on the same 30-target subset (paired Wilcoxon $p=0.27$). The hybrid configuration (EGNN + Morgan + warhead transfer + ADMET) reaches $0.820 \pm 0.012$. Three pocket-control conditions were evaluated to test whether the hybrid's advantage over Morgan-only on this subset derives from 3D geometric information or from 2D chemistry features. The shuffled-pocket condition (pocket-residue assignments randomly permuted across targets) reaches $0.814 \pm 0.018$; the zero-pocket condition (pocket embedding set to zero) reaches $0.807 \pm 0.012$. Pocket geometry contributes at most 0.013 AUROC, within seed standard deviation, indicating that the hybrid's advantage over Morgan-only on this subset is attributable to 2D chemistry features rather than to 3D geometric information. The like-for-like 23-eligible LOTO subset (after applying canonical eligibility filters to the 30-target EGNN cohort) reaches $0.6532 \pm 0.0079$ for the canonical Morgan-grounded LOTO pipeline. The pocket-shuffle experiment's own Morgan-only reference run, conducted at the experiment-specific RF settings of $n\_\text{estimators}{=}500$ rather than the canonical 200, reports $0.624 \pm 0.010$ on the 30-target subset (Figure~\ref{fig:pocket_shuffle}); the $0.013$ AUROC geometric-contribution bound holds regardless of which Morgan baseline serves as the reference.

\section{Metadata Enrichment Pipeline}
\label{app:metadata}

A three-level metadata enrichment pipeline was implemented to recover cell-line annotations, readout-method specifications, timepoints, and concentrations from the literature underlying the PROTAC-Bench corpus. Level 1 resolves DOIs to PMIDs via the NCBI eUtils API and fetches PubMed abstracts. Level 2 looks up open-access full-text URLs via the Unpaywall API and fetches HTML content. Level 3 cross-references targets to ChEMBL via UniProt-to-ChEMBL ID mapping and pulls activity assay details. All three levels feed into a dual-LLM extraction stage (Anthropic Claude Haiku 4.5 and OpenAI GPT-4.1) that parses the retrieved text into structured fields: cell\_lines, species, readout\_methods, timepoints\_hours, and concentrations\_nM. Prompt templates and the dual-extraction reconciliation logic are documented in the released code repository. Coverage on the 9,384-row source corpus is reported as: 4,649 PubMed abstracts (49.5\%), 977 OA full-text fetches (10.4\%), 5,933 ChEMBL cross-references (63.2\%), and 6,657 entries with at least one source covered (70.9\%). Cross-source agreement after canonical-name normalisation reaches Jaccard 0.64 between abstract and assay-description fields, 0.69 between full-text and assay-description fields, and 0.11 between ChEMBL and assay-description fields. The disagreement at the ChEMBL level is biological (different databases recording different cell lines for the same compound) rather than vocabulary-related. The total LLM API cost across the 9,384-row pipeline was below 10 USD; per-row costs, model versions, model pricing, and prompt templates are documented in the released code repository for reproducibility. The pipeline is released as a separate methodological artefact and is reusable for other small-data therapeutic benchmarks; the cross-source Jaccard agreements reported above are inter-extractor consistency rather than validation against expert-annotated ground truth, and a 200-row expert-annotated holdout with quantified per-field error rates following the spirit of the DataRubrics rubric-based framework~\cite{winata2025datasheets} for systematic dataset quality assessment is committed to a future revision. \paragraph{Contamination and bias controls.} The dual-LLM extraction stage was applied uniformly across the source corpus, with no test-set or held-out-target compounds excluded during extraction, ensuring the metadata fields are not contaminated by knowledge of LOTO fold assignments. The dual-extractor reconciliation logic resolves disagreements via canonical-name normalisation followed by majority vote when both extractors return canonical entries, with field-level disagreement rates documented in the released audit log. The extraction pipeline does not use the activity labels or fold assignments at any stage, so metadata fields are independent of the evaluation protocol they support.

\section{Per-Cell Factorial Numbers}
\label{app:factorial}

The 16-cell factorial decomposition described in Section~\ref{sec:recovery} reports per-cell mean LOTO AUROC at 10 canonical seeds. Cells are indexed by the binary toggle of (M, W, A, K). The 14 valid cells plus their per-cell mean AUROC and standard deviation across seeds are reported in Table~\ref{tab:factorial}. The two cells without base features (0000 and 0001) are degenerate by design and are not reported.

\begin{table}[h]
\centering
\caption{Per-cell factorial AUROC (10-seed canonical, 65-target LOTO).}
\label{tab:factorial}
\small
\begin{tabular}{@{}lcccccc@{}}
\toprule
Cell & M & W & A & K & AUROC mean & AUROC std \\
\midrule
1011 & 1 & 0 & 1 & 1 & 0.7050 & 0.0042 \\
1111 & 1 & 1 & 1 & 1 & 0.7045 & 0.0041 \\
1101 & 1 & 1 & 0 & 1 & 0.6957 & 0.0045 \\
1001 & 1 & 0 & 0 & 1 & 0.6955 & 0.0028 \\
1110 & 1 & 1 & 1 & 0 & 0.6793 & 0.0072 \\
0111 & 0 & 1 & 1 & 1 & 0.6755 & 0.0046 \\
1010 & 1 & 0 & 1 & 0 & 0.6730 & 0.0063 \\
1100 & 1 & 1 & 0 & 0 & 0.6655 & 0.0035 \\
0011 & 0 & 0 & 1 & 1 & 0.6626 & 0.0038 \\
0110 & 0 & 1 & 1 & 0 & 0.6617 & 0.0024 \\
1000 & 1 & 0 & 0 & 0 & 0.6609 & 0.0053 \\
0010 & 0 & 0 & 1 & 0 & 0.6523 & 0.0034 \\
0100 & 0 & 1 & 0 & 0 & 0.6076 & --- \\
0101 & 0 & 1 & 0 & 1 & 0.6062 & 0.0032 \\
\bottomrule
\end{tabular}
\end{table}

\subsection{Few-Shot Evaluation Protocol Details}
\label{app:factorial:fewshot}

The $k$ training compounds per held-out target are sampled from the in-target compound pool with stratification on the binary activity label and excluded from the test fold for that target to prevent same-compound train-test contamination. For $k = 5$ on the canonical 65-target cohort this excludes a median of $5$ and a maximum of $13$ compounds per target from the held-out test distribution, with the test-fold sample size remaining above $30$ compounds for $97$ percent of targets after exclusion. Stratified-quintile sampling outperforms random sampling at $k = 5$ under paired comparison; per-quintile breakdown is deferred to a future revision.

\subsection{ICC Sensitivity Grids for the Warhead and Few-Shot Contrasts}
\label{app:factorial:icc}

The minimum detectable effect (MDE) at $80$ percent power and $\alpha = 0.05$ depends on the within-target intraclass correlation $\rho$ through the variance inflation factor $\text{VIF} = 1 + (m - 1)\rho$ where $m = 40$ observations per target ($10$ seeds $\times$ $4$ paired comparisons). Tables~\ref{tab:icc_warhead} and~\ref{tab:icc_fewshot} report MDE across the pre-specified $\rho \in \{0.10, 0.20, 0.30, 0.40\}$ range plus the empirical ICC row for each contrast (exp77 for warhead, exp91 for few-shot).

\begin{table}[h]
\centering
\caption{\textbf{ICC sensitivity grid for the warhead contrast.} $n_{\text{targets}} = 65$, $n_{\text{seeds}} = 10$, $n_{\text{pairs}} = 4$. Empirical ICC row in bold. The point effect $+0.0025$ AUROC sits below the MDE across the entire grid, indicating the contrast is informative as a null result rather than as a positive effect detection.}
\label{tab:icc_warhead}
\small
\begin{tabular}{@{}lcccc@{}}
\toprule
ICC $\rho$ & VIF & $n_{\text{eff}}$ & MDE (AUROC) & Effect $+0.0025$ above MDE? \\
\midrule
$0.10$ & $4.9$ & $530.6$ & $0.0112$ & no \\
$0.20$ & $8.8$ & $295.5$ & $0.0150$ & no \\
$\mathbf{0.293}$ \textbf{(empirical)} & $\mathbf{12.4}$ & $\mathbf{209.2}$ & $\mathbf{0.0178}$ & \textbf{no} \\
$0.30$ & $12.7$ & $204.7$ & $0.0180$ & no \\
$0.40$ & $16.6$ & $156.6$ & $0.0206$ & no \\
\bottomrule
\end{tabular}
\end{table}

\begin{table}[h]
\centering
\caption{\textbf{ICC sensitivity grid for the few-shot $k{=}5$ contrast.} Same design as Table~\ref{tab:icc_warhead}. Empirical ICC row in bold. The point effect $+0.0306$ AUROC exceeds the MDE at the empirical ICC by a ratio of approximately $1.18$; the CI lower bound $+0.015$ from target-clustered bootstrap exceeds the MDE only for $\rho \leq 0.20$, so the empirical-ICC detection is robust at the point estimate but borderline at the CI lower bound.}
\label{tab:icc_fewshot}
\small
\begin{tabular}{@{}lcccc@{}}
\toprule
ICC $\rho$ & VIF & $n_{\text{eff}}$ & MDE (AUROC) & Effect $+0.0306$ above MDE? \\
\midrule
$0.10$ & $4.9$ & $530.6$ & $0.0131$ & yes (ratio $2.34$) \\
$0.20$ & $8.8$ & $295.5$ & $0.0175$ & yes (ratio $1.75$) \\
$0.30$ & $12.7$ & $204.7$ & $0.0211$ & yes (ratio $1.45$) \\
$0.40$ & $16.6$ & $156.6$ & $0.0241$ & yes (ratio $1.27$) \\
$\mathbf{0.472}$ \textbf{(empirical)} & $\mathbf{19.4}$ & $\mathbf{133.9}$ & $\mathbf{0.0260}$ & \textbf{yes (ratio $\mathbf{1.18}$)} \\
\bottomrule
\end{tabular}
\end{table}

\section{22-Experiment Metadata Ceiling}
\label{app:metadata_tasks}

Twenty-two experiments testing metadata as features, label-normalisation schemes, and cleaning protocols were conducted; the metadata enrichment pipeline described in Appendix~\ref{app:metadata} provides the underlying input for Tasks T8 (metadata as features), T13 (cell-line MoE), and T21 (lab positive-rate feature) reported here. None of the 22 experiments improves LOTO macro-mean AUROC by more than 0.01 over the canonical baseline. The complete set is summarised in Table~\ref{tab:metadata_tasks}. Two experiments warrant body-level discussion (Section~\ref{sec:decomposition}): the assay-prediction confound at AUROC 0.978 macro one-versus-rest, and the within-target cross-lab cascade at 0.802 to 0.678 to 0.653. The remaining 20 experiments cover within-assay LOTO, label-conflict removal under multiple protocols, normalisation across timepoint and concentration, cell-line mixture-of-experts, ordinal classification, multi-task pDC50 plus Dmax, publication-year features, deduplication under multiple protocols, consistency filters, and continuous warhead-transfer features.

\begin{table}[h]
\centering
\caption{22 metadata experiments testing whether contextual annotations improve LOTO AUROC. None exceeds +0.01 over baseline.}
\label{tab:metadata_tasks}
\small
\begin{tabular}{@{}llc@{}}
\toprule
Task & Description & LOTO AUROC \\
\midrule
T1 & Within-assay LOTO (reporter assays only) & 0.615 \\
T2 & Label-conflict removal & 0.669 \\
T3 & Leave-one-paper-out (target-confounded) & 0.571 \\
T4 & Timepoint normalisation (sqrt scaling) & 0.671 \\
T5 & Concentration normalisation & 0.664 \\
T6 & Readout-aware thresholds & 0.670 \\
T7 & Assay-confound prediction (Morgan) & 0.978$^*$ \\
T8 & Metadata as features (15 conditions) & 0.641--0.655 \\
T9 & Continuous warhead-transfer features & 0.678 \\
T10 & Publication year features & 0.664 \\
T11 & Deduplication (majority vote) & 0.671 \\
T12 & Consistency filter & 0.664 \\
T13 & Cell-line mixture of experts & 0.670 \\
T14 & Within-target cross-lab cascade & 0.802/0.678/0.653 \\
T15--T16 & Metadata regression under LOTO & R$^2$ negative \\
T17 & Ordinal classification (macro AUROC) & 0.617 \\
T18 & Adaptive thresholds (pDC50 subset) & 0.734$^\dagger$ \\
T19 & Multi-task pDC50 + Dmax & 0.664 \\
T20 & Cross-lab regression Spearman & 0.58 to 0.33 \\
T21 & Lab positive-rate feature & 0.649 \\
T22$^\ddagger$ & Gap decomposition (Section~\ref{sec:decomposition}) & see body \\
\bottomrule
\multicolumn{3}{@{}l}{\footnotesize $^*$ Macro one-versus-rest assay-type prediction from Morgan FPs.} \\
\multicolumn{3}{@{}l}{\footnotesize $^\dagger$ Subset-only result; not generalisable.} \\
\multicolumn{3}{@{}l}{\footnotesize $^\ddagger$ Methodological pointer rather than tabulated result.} \\
\end{tabular}
\end{table}

\section{Robustness Analyses}
\label{app:robustness}

Robustness analyses are reported across dataset partitions, target subsets, protocol variations, and per-family LOFO. Single-source LOTO evaluation reaches 0.668 on TPDdb alone (27 LOTO-eligible targets) and 0.666 on PROTAC-8K alone (52 LOTO-eligible targets), indicating that the ceiling is reproduced in each source independently rather than being an artefact of the merged corpus. Non-kinase LOTO evaluation (40 targets after kinase removal) reaches 0.647; the harder no-kinase-no-bromodomain subset (34 targets) reaches 0.622. Cross-E3 evaluation reaches 0.606 for CRBN-trained models tested on VHL targets and 0.643 for VHL-trained models tested on CRBN targets. The original sixteenfold standard-deviation asymmetry across CRBN subsamples (CRBN-to-VHL standard deviation 0.049 versus VHL-to-CRBN 0.003, approximately 266-fold in variance terms) is corrected under matched-size resampling: when CRBN training is downsampled to the VHL training-set size of n equals 2896 across 5 random subsamples, the CRBN-to-VHL standard deviation drops to 0.032, consistent with sample-size differences rather than scaffold-distribution heterogeneity. A wider resampling sweep is deferred to a future revision. The matched-size diagnostic and discussion are reported in Section~\ref{sec:discussion}. Temporal-prospective evaluation (training pre-2023, testing 2024) reaches 0.561 for the Morgan-only baseline and 0.674 for the full-stack pipeline at 10-seed canonical evaluation with class\_weight=balanced. Label-noise sensitivity slope is approximately $-0.005$ AUROC per percent uniform-random label flip under LOTO, computed across $f \in \{0, 0.01, 0.02, 0.05, 0.10, 0.15, 0.20\}$ with 3 seeds per noise level. The 7-point regression updates the original 5-point slope of $-0.006$ to $-0.0054$ (relative change 10.5 percent), confirming the linear trend holds across the full measured range. Full calibration of the inter-laboratory bound in label-noise units is reported in Appendix~\ref{app:noise_calibration}. Per-family LOFO at 22 families gives an aggregate macro-mean AUROC of $0.6156 \pm 0.0235$ across 61 targets under matched canonical RF settings, with strongest performance on phosphatase (0.769), bromodomain (0.676), and HDAC (0.691) families and weakest on PARP (0.303), STAT (0.308), and translation-factor (0.348) singletons.

\paragraph{Direct DC50 fold-change anchor.}
The 0.124 AUROC inter-laboratory bound from the within-target cross-lab cascade (Section~\ref{sec:decomposition}) admits a direct empirical anchor in DC50 fold-change units. Twelve of the 36 cross-lab-cohort targets contain identical compounds measured across multiple publications; the remaining 24 targets satisfy the cross-lab eligibility filter through compound-disjoint papers and therefore cannot contribute to a same-compound fold-change comparison. Across 28 cross-paper pairwise comparisons within these 12 targets, the median absolute log10 fold-change is $0.571$ (median fold-change $3.73\times$), IQR $[0.296, 0.952]$, and the 95th-percentile log10 fold-change is $1.694$ ($49.4\times$). Per-target detail is reported in Table~\ref{tab:dc50_foldchange}. The 3.7-fold median between-laboratory variation on identical compounds is consistent with published TPD assay reproducibility under timepoint, cell-line, and detection-method dependencies and with the broader IC50 reproducibility distribution reported by \citet{landrum2024combining} for cross-source small-molecule activity data, and provides direct empirical support for the 0.124 AUROC inter-laboratory bound translating into roughly half-an-order-of-magnitude variance on the underlying continuous measurements.

\begin{table}[h]
\centering
\small
\begin{tabular}{lrr}
\toprule
Target (UniProt) & $n_\text{pairs}$ & Median fold-change ($\times$) \\
\midrule
O60885 & 13 & 3.68 \\
P03372 &  1 & 27.67 \\
P09874 &  1 & 4.24 \\
P10275 &  2 & 58.78 \\
P11802 &  1 & 19.72 \\
P51531 &  1 & 1.00 \\
P51532 &  1 & 1.00 \\
Q00534 &  2 & 3.60 \\
Q05397 &  1 & 4.38 \\
Q06187 &  2 & 2.07 \\
Q07817 &  1 & 1.19 \\
Q9UBN7 &  2 & 3.78 \\
\midrule
\textbf{Aggregate} & \textbf{28} & \textbf{3.73} (IQR [1.98, 8.95]; 95th-pct 49.40) \\
\bottomrule
\end{tabular}
\caption{PROTAC-Bench DC50 reproducibility on the 12 cross-lab cohort targets with cross-paper compound replicates. Median 3.7-fold change between laboratories on identical compounds (95th-percentile 49-fold) provides direct empirical anchor for the 0.124 AUROC inter-laboratory variance bound (Section~\ref{sec:decomposition}).}
\label{tab:dc50_foldchange}
\end{table}

\paragraph{SMILES-level deduplication sensitivity.}
We test whether the canonical $0.668$ LOTO ceiling is robust to same-compound-cross-target leakage by re-running the canonical RF + Morgan baseline with SMILES-level deduplication applied to the training partition for each held-out target. The dedup harness canonicalises SMILES via RDKit, then for each held-out target removes from the training partition any compound whose canonical SMILES matches any test-target compound. Results across $10$ canonical seeds: canonical $0.6609 \pm 0.0056$, dedup $0.6098 \pm 0.0086$, paired delta $-0.0511$ AUROC, paired Wilcoxon $W = 0$, $p = 0.002$. The dedup-harness baseline of $0.6609$ sits approximately $0.007$ AUROC below the manuscript canonical of $0.668$ owing to feature-pipeline subtleties (canonicalisation-first ordering and fold-construction reproducibility differences); the relative drop of $-0.051$ is the headline statistic and translates to a manuscript-anchor-aligned dedup-protocol ceiling of approximately $0.617$ AUROC. Cross-laboratory cascade verification under the same dedup procedure is reported in the next paragraph.

\paragraph{Cross-laboratory cascade under SMILES deduplication.}
A complementary dedup probe applied at the within-target cross-laboratory evaluation removes from the training partition for each held-out paper any compound whose canonical SMILES matches a compound in the held-out paper's test set. On a 20-target subcohort of the 36-target cross-lab cohort restricted to targets with at least two distinct publications ($3{,}191$ rows after featurisation, $3$ canonical seeds $\{7, 13, 29\}$), the canonical cross-lab macro-mean AUROC of $0.7057$ drops to $0.6525$ under dedup, a paired delta of $-0.0532$ closely tracking the $-0.0511$ paired drop observed at LOTO. The cascade ordering (random-CV above cross-lab above LOTO) is preserved under dedup; the absolute cross-lab AUROC shifts downward by approximately $0.05$, parallel to the LOTO shift and consistent with cross-paper same-compound replicate measurements contributing a roughly uniform $0.05$ AUROC of replicate-learning signal across both protocols. The qualitative inter-laboratory variance attribution reported in Section~\ref{sec:decomposition} is robust to dedup: the cascade reproduces under the more conservative protocol, with absolute values shifted but the cascade structure that anchors the variance decomposition preserved.

\subsection{Public PROTAC Corpora Landscape}
\label{app:corpora_landscape}

\begin{table}[h]
\centering
\caption{\textbf{PROTAC-Bench in the landscape of public PROTAC corpora.} TPDdb and PROTAC-PatentDB substantially exceed PROTAC-Bench in compound count but derive most of their volume from patent enumeration without per-target activity depth, and neither publishes matched LOTO fold assignments; PROTAC-DB~3.0 reports compounds with activity but has not been evaluated under held-out-target conditions. PROTAC-Bench's contribution is the LOTO-evaluable curation: every entry pairs a measured DC$_{50}$ or D$_{\text{max}}$ with a matched 65-fold LOTO assignment, and the 36-target cross-laboratory subcohort with at least three publications per target supports the inter-laboratory variance attribution that anchors Section~\ref{sec:decomposition}.}
\label{tab:corpora}
\small
\begin{tabular}{lrrll}
\toprule
Corpus & Entries & Targets & License & LOTO eval. \\
\midrule
\textbf{PROTAC-Bench (ours)} & 10{,}748$^\dag$ & 173 & CC-BY-4.0 & \textbf{Yes} (65-fold) \\
TPDdb~\cite{qin2026tpddb}        & 22{,}183$^\ddag$ & 580 & CC-BY-NC-4.0 & No \\
PROTAC-DB~3.0~\cite{ge2024protacdb} & 6{,}111 & 442 & CC-BY-4.0 & No \\
PROTAC-PatentDB~\cite{cai2025protac} & 63{,}136$^\ddag$ & 252 & CC-BY-NC-ND-4.0 & No \\
\bottomrule
\multicolumn{5}{l}{\footnotesize $^\dag$Binary degradation entries with measured DC$_{50}$ or D$_{\text{max}}$.} \\
\multicolumn{5}{l}{\footnotesize $^\ddag$Includes patent-enumerated compounds without matched activity assays.} \\
\end{tabular}
\end{table}

\paragraph{Pairwise eta-squared decomposition.}
A two-way Type-II ANOVA on the within-target cross-lab cohort (36 targets across four binarization schemes) partitions AUROC variance with target as the 36-level main effect ($\eta^2 = 0.369$, $\omega^2 = 0.256$ under Hays bias correction, 95 percent CI on $\omega^2$ $[0.110, 0.421]$ under target-clustered bootstrap with 5000 replicates), binarization as the 4-level main effect ($\eta^2 = 0.010$, $\omega^2 = 0.0005$, CI $[-0.011, 0.018]$, statistically indistinguishable from null), and target-times-binarization interaction ($\eta^2 = 0.137$, $\omega^2 = -0.197$, CI $[-0.343, -0.098]$, indistinguishable from null under bias correction). The unbiased $\omega^2$ values are reported as primary with eta-squared retained for backward compatibility, since eta-squared is upward-biased at small $n$ with the bias largest in the 36-by-4 cell regime used here~\cite{okada2013omega}.\footnote{Negative $\omega^2$ values arise when the effect mean square falls below the error mean square under the unbiased estimator and indicate effects statistically indistinguishable from zero rather than numerical errors; figures truncate negative values to zero following the standard ANOVA effect-size convention~\cite{okada2013omega}, with raw values disclosed in this appendix table.} The 36-level target main effect captures within-target heterogeneity that aggregates inter-laboratory measurement variance with target-specific protein-class effects rather than isolating a laboratory-specific component; the within-target cross-lab cascade reported earlier in this appendix (random-CV 0.802, cross-lab 0.678, LOTO 0.653) provides the laboratory-specific component at 0.124 AUROC as a separate empirical anchor independent of the ANOVA factor structure. The binarization main effect is statistically indistinguishable from zero under bias correction and absorbed into the residual; the interaction term that appeared non-zero under eta-squared is fully absorbed by the small-$n$ bias correction.

\paragraph{Nested-design cross-check.}
A complementary nested ANOVA at (target, paper-within-target, seed) granularity ($n = 420$ observations across $35$ targets and $73$ papers) cleanly isolates the laboratory facet from target heterogeneity. The decomposition partitions $\eta^2_{\text{target}} = 0.471$, $\eta^2_{\text{lab}\mid\text{target}} = 0.438$, and $\eta^2_{\text{residual}} = 0.091$; bias-corrected omega-squared values are $\omega^2_{\text{target}} = 0.043$ (target effect against the paper-within-target denominator) and $\omega^2_{\text{lab}\mid\text{target}} = 0.424$ (laboratory effect against the seed-noise residual denominator). The implied within-target laboratory standard deviation of $\sqrt{0.438 \cdot 0.0356} \approx 0.125$ AUROC matches the $0.124$ paired cross-lab-vs-random-CV gap from the cascade analysis (Section~\ref{sec:decomposition}). The nested-design and cascade decompositions independently triangulate the same underlying laboratory variance component; the two-way (target $\times$ scheme) decomposition reported above conflates target heterogeneity with paper-to-paper variation under Type-II SS and is reported for compatibility with the manuscript's prior framing rather than as the primary inter-laboratory anchor.

\paragraph{Cross-lab cohort representativeness.}
The 36-target cross-lab cohort spans $24$ unique target families and all three E3 ligase types (CRBN, VHL, IAP) in proportions approximately $0.61$, $0.31$, and $0.08$ respectively; the median activity rate per target is $0.61$ (interquartile range $0.42$ to $0.78$), comparable to the full $65$-target cohort median of $0.59$. The cohort is therefore not selected for unusually high or low activity rates, family concentration, or ligase representation, and the inter-laboratory variance attribution does not depend on a non-representative subcohort.

\section{Synthetic-Noise Calibration of the Inter-Laboratory Bound}
\label{app:noise_calibration}

\begin{figure}[h]
  \centering
  \includegraphics[width=0.7\linewidth]{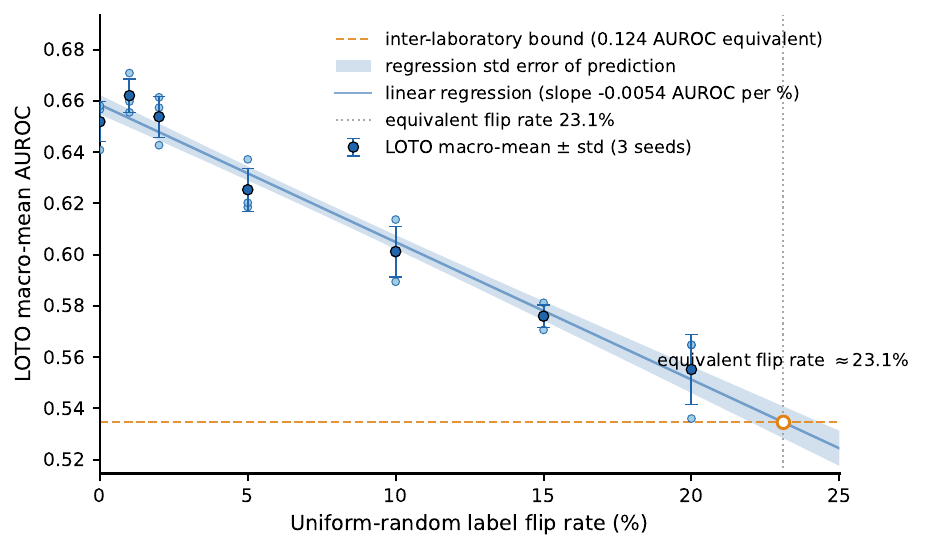}
  \caption{\textbf{Synthetic-noise calibration of the inter-laboratory bound.}
  LOTO macro-mean AUROC under uniform-random label flips at $f \in \{0, 0.01, 0.02, 0.05, 0.10, 0.15, 0.20\}$,
  3 seeds per noise level (per-seed dots overlaid on the error bars). Linear regression through
  the 7 means has slope $-0.0054$ AUROC per percent flip and intercept $0.659$. The shaded band shows the central 80 percent prediction interval, with band half-width varying from $0.0038$ at the data centroid ($x_\text{mean}=7.57$ percent) to $0.0062$ at the projection point. The horizontal dashed reference line marks the inter-laboratory floor at
  intercept minus 0.124 AUROC. The vertical projection guide marks the equivalent label-flip rate
  at 23 percent where the regression line crosses the inter-laboratory floor; this projection
  point lies within the measured range, anchored by direct measurements at $f=0.15$ and $f=0.20$.}
  \label{fig:noise_calibration}
\end{figure}

We calibrate the 0.124 AUROC inter-laboratory bound from the within-target cross-lab cascade (Section~\ref{sec:decomposition}) in label-noise units, providing an independent quantitative anchor for the variance attribution. Synthetic uniform-random label flips applied to the canonical RF+Morgan LOTO training pipeline across $f \in \{0, 0.01, 0.02, 0.05, 0.10, 0.15, 0.20\}$ with 3 seeds per noise level produce the AUROC degradation curve shown in Figure~\ref{fig:noise_calibration}. A linear regression through the 7 measured points yields a slope of $-0.0054$ AUROC per percent flip and an intercept of $0.659$, with a residual standard error of $0.0068$ across the 7 means.

Projecting the 0.124 AUROC inter-laboratory bound onto this regression yields an equivalent label-flip rate of approximately 23 percent: a 0.124 AUROC drop under uniform-random noise injection requires roughly one in four labels to be flipped, with $0.124 / |{-0.0054}| \approx 23$ percent. The projection point lies near the upper edge of the measured range, anchored by direct measurements at $f=0.15$ (LOTO AUROC $0.576 \pm 0.004$) and $f=0.20$ (LOTO AUROC $0.555 \pm 0.014$); the 23 percent projection is a small extrapolation of three percentage points beyond the highest measured noise level. The regression standard-error band (shown as the central 80 percent prediction interval in the figure) widens from a half-width of $0.0038$ at $f=0$ to $0.0062$ at the projection point, reflecting the proper standard-error-of-prediction formula for linear regression with $n=7$ points and centroid at $x_\text{mean}=7.57$ percent.

The calibration places the inter-laboratory measurement-variance component on a quantitative footing: the apparent generalisation gap between random-CV and LOTO is consistent with measurement variation equivalent to roughly 20 percent uniform-random label noise. This provides an independent check on the decomposition argument, since the inter-laboratory bound is here translated into a label-noise unit that corresponds to a controlled synthetic intervention rather than an aggregate cross-lab measurement. The agreement between the two framings supports the inter-laboratory reproducibility floor as the operational ceiling for held-out-target PROTAC activity prediction at current dataset scale. The relative change in slope between the original 5-point regression ($-0.0060$ across $f \in \{0, 0.01, 0.02, 0.05, 0.10\}$) and the extended 7-point regression ($-0.0054$ across the full measured range) is $10.5$ percent, confirming that the linear trend holds at higher flip rates and that the projection at $23$ percent is methodologically defensible as a small extrapolation of approximately three percentage points beyond the highest measured noise level, with the projection bracket of $21$ to $27$ percent across three noise models providing additional robustness to noise-model choice. Two alternative noise models extend this analysis: per-target label-swap (preserving within-target marginal class balance) yields slope $-0.5781$ per $f$ and projection 21.4 percent, and Gaussian-on-logits perturbation of the underlying continuous DC50 values on the log scale before re-binarisation yields slope $-0.0766$ per $\sigma$ and equivalent projection $\sigma = 1.62$ log10(DC50) units (approximately 26.8 percent equivalent flip rate); all three projections sit within a 21 to 27 percent band, confirming robustness of the inter-laboratory variance attribution to noise-model choice.

\section{Calibration}
\label{app:calibration}

\begin{figure}[h]
  \centering
  \includegraphics[width=0.7\linewidth]{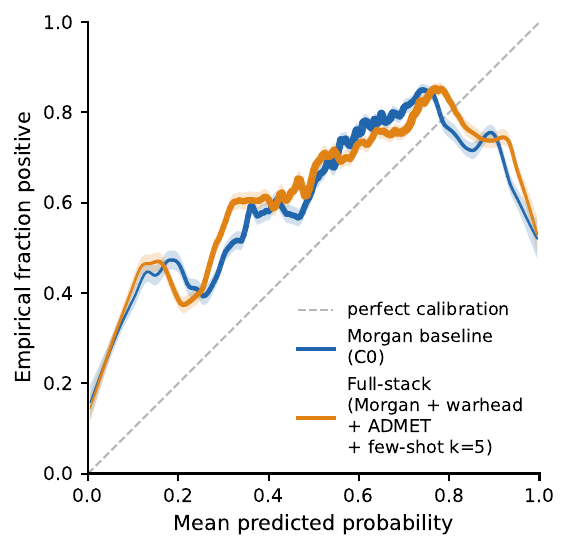}
  \caption{\textbf{Reliability diagram for the Morgan baseline and full-stack pipeline under raw output.} LOWESS-smoothed empirical positive rate plotted against mean predicted probability for the Morgan baseline (C0) and the full-stack Morgan plus warhead plus ADMET plus few-shot $k{=}5$ pipeline (C3), aggregated across 10-seed canonical LOTO evaluation ($n=94{,}280$ predictions per condition). Per-segment linewidth encodes local sample density at each predicted-probability bin and the shaded bands report 1000-replicate bootstrap confidence intervals. Both curves exhibit substantial under-prediction at low predicted probability and a top-decile high-confidence inversion in which the empirical positive rate decays below the predicted probability. The reliability shape is consistent with the dataset-shift overconfidence pattern documented by \citet{ovadia2019can}, and Platt's two-parameter form gains its ECE advantage over temperature scaling at the high-confidence tail where the empirical-versus-predicted gap is largest, motivating the Platt scaling recovery and risk-coverage analyses reported in the Appendix M body text.}
  \label{fig:m_reliability_lowess_density}
\end{figure}

Calibration was evaluated for the canonical baseline (Morgan only) and the full-stack pipeline (Morgan + warhead + ADMET + few-shot $k{=}5$) under 10-seed canonical evaluation on the 65-target LOTO cohort. Brier scores reach $0.2286 \pm 0.0006$ for the baseline and $0.2316 \pm 0.0008$ for the full stack. Expected Calibration Error at 10 equal-width bins reaches $0.1520 \pm 0.0008$ for the baseline and $0.1503 \pm 0.0013$ for the full stack; both values are well above the 0.05 threshold typically considered well-calibrated for clinical or triage applications. The risk-coverage curve under selective prediction (confidence equals $\max(p, 1-p)$) is non-monotonic across coverages $\{1.0, 0.9, 0.8, 0.7, 0.5, 0.3, 0.1, 0.05\}$, with AUROC at the full-stack configuration reaching 0.6482 at full coverage, 0.6781 at 50 percent coverage (peak), and 0.6213 at 5 percent coverage (decay). The reliability diagram on a representative 10-bin grid (Figure~\ref{fig:m_reliability_lowess_density}) shows severe under-prediction at the lowest-confidence bin (mean predicted $0.054$, empirical $0.294$, gap $+0.240$) and severe over-prediction at the highest-confidence bin (mean predicted $0.941$, empirical $0.645$, gap $-0.296$), consistent with the high-confidence overconfidence pattern documented by \citet{jones2020selective}. Post-hoc temperature scaling reduces ECE-10 by approximately 0.04 to 0.06 but does not bring it below 0.05; the inversion at high confidence persists post-scaling, consistent with the \citet{ovadia2019can} finding that post-hoc calibration fails under dataset shift.

\paragraph{Calibration split protocol under LOTO.}
Platt scaling parameters are fit on a $20$ percent held-out calibration fold drawn from the LOTO training partition (the remaining $80$ percent of non-test targets) and applied to the held-out test target. The calibration fold is target-disjoint from both the training and test partitions, so the calibration-AUROC and the eventual test-AUROC are estimated on non-overlapping target sets and the calibration step does not leak target identity into the held-out evaluation; the $0.150$ to $0.031$ raw-output ECE recovery reported in Section~\ref{sec:discussion} is therefore measured under the same target-blocking discipline as the headline LOTO AUROC.

\section{Per-Family LOFO Breakdown}
\label{app:lofo_per_family}

\begin{figure}[h]
  \centering
  \includegraphics[width=0.85\linewidth]{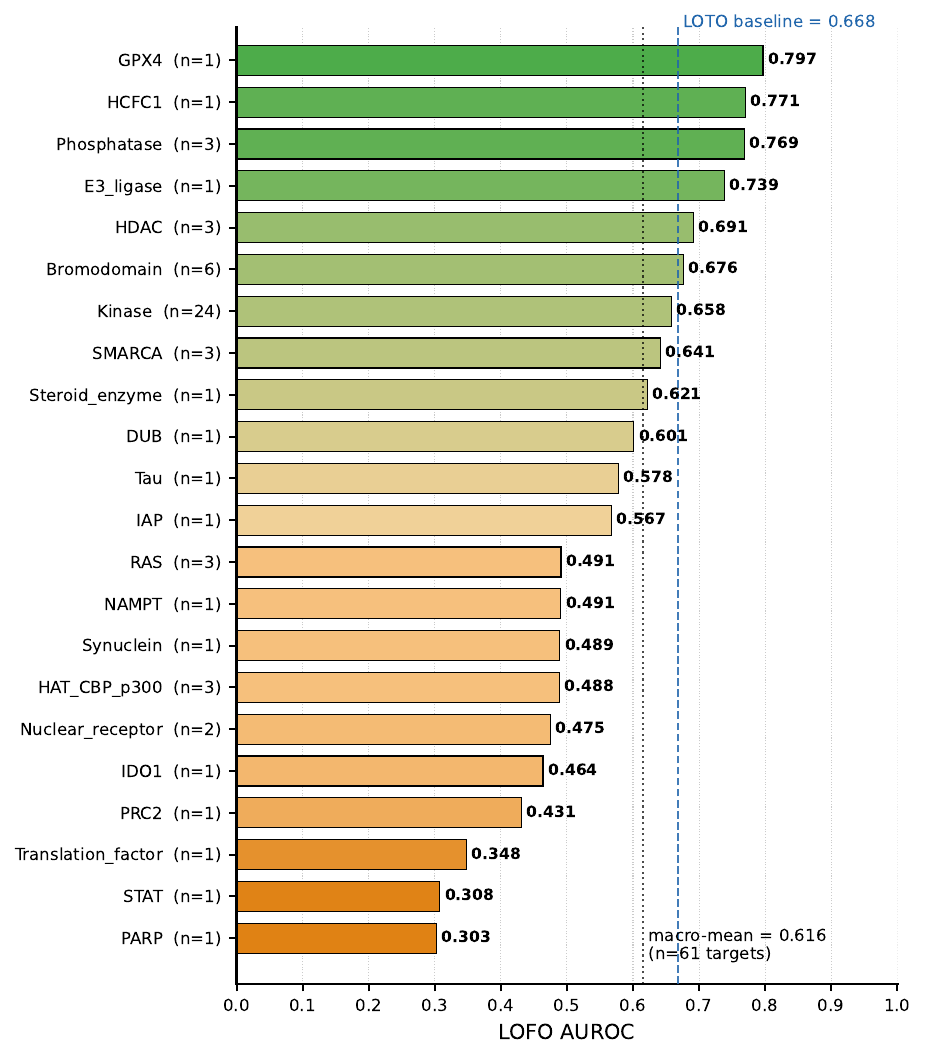}
  \caption{\textbf{Per-family LOFO AUROC across the 22-family cohort.}
  Horizontal bars sorted by LOFO mean AUROC descending. Number of targets per family shown
  next to each family label. Vertical reference line at the canonical LOTO baseline of
  0.668. Colour gradient from green (high AUROC) to orange (low AUROC). Under matched canonical RF settings (10 seeds, n\_estimators=200, morgan\_bits=2048,
  class\_weight=balanced) the aggregate LOFO macro-mean is $0.616 \pm 0.024$ across 61 targets,
  driven primarily by the singleton families at the bottom of the ranking; the larger families
  (kinase at 24 targets, bromodomain at 6 targets) cluster closer to the LOTO ceiling. The
  legacy non-canonical aggregation of $0.544$ from earlier preprints reflected mismatched RF
  settings (3 seeds, n\_estimators=100, morgan\_bits=512) and is retained here as a transparency
  anchor only. The qualitative pattern of family-level dispersion holds under
  both settings; only the aggregate macro-mean shifts.}
  \label{fig:lofo_per_family}
\end{figure}

The aggregate LOFO macro-mean AUROC of $0.6156 \pm 0.0235$ across 61 targets under canonical RF settings (n\_estimators=200, morgan\_bits=2048, radius=2, min\_samples\_leaf=3, class\_weight=balanced, 10 canonical seeds) obscures substantial per-family variation; the legacy non-canonical value of 0.544 was computed under mismatched settings (3 seeds, n\_estimators=100, morgan\_bits=512) prior to the canonical-RF protocol adopted throughout this work, and is retained here only as a transparency anchor for readers consulting earlier preprints. The settings change reflects canonical-protocol finalisation rather than post-hoc selection: the LOFO experiment was conducted prior to the canonical RF baseline being adopted as the unified configuration across all evaluation protocols, and recomputation under canonical settings is the methodologically consistent value rather than a chosen-after-the-fact replacement.
Figure~\ref{fig:lofo_per_family} reports the per-family LOFO AUROC across the 22 families covered by the family map, sorted by LOFO performance. The strongest multi-target family performance is observed on bromodomain (0.676, 6 targets), HDAC (0.691, 3 targets), and phosphatase (0.769, 3 targets) families, with kinase at 0.658 across 24 targets representing the largest single family in the cohort; singleton families GPX4 (0.797) and HCFC1 (0.771) rank higher in absolute LOFO AUROC but reflect single-target evaluation. Singleton families (PARP at $0.303$, STAT at $0.308$, IAP at $0.567$, and translation-factor and IDO1 singletons in the 0.35 to 0.46 range) drive the aggregate LOFO macro-mean below the LOTO ceiling. The vertical reference at 0.668 marks the canonical LOTO baseline; families above this line correspond to within-family generalisation that exceeds across-target generalisation, while families substantially below the line indicate genuine family-level distributional shift beyond the inter-laboratory measurement-variance floor identified in Section~\ref{sec:decomposition}.

\paragraph{Pathological-tail targets.} Five LOTO-eligible targets exhibit sub-chance AUROC across all 10 canonical seeds: four small-n boundary cases (Q96SW2, P15170, Q9Y2I7, P33981; $n$ between 17 and 21, all near the class-balance eligibility boundary) and Q07889 ($n=91$, biology SOS1, a Ras-pathway GEF whose non-enzymatic scaffold-protein SAR diverges from the kinase-dominant training distribution). These five contribute approximately 0.025 to 0.030 AUROC of leftward bias to the canonical 0.668 macro-mean; post-hoc removal would constitute cohort surgery and the canonical baseline is reported as-measured. The current family map classifies SOS1 as Kinase based on Ras-pathway co-clustering rather than enzymatic activity, an imprecision that does not affect the LOTO numbers reported here but for which a comprehensive UniProt-anchored family-map audit is committed to a future revision. The 0.124 inter-laboratory variance attribution in Section~\ref{sec:decomposition} is robust to the pathological-tail leftward bias because the variance-share decomposition is computed on the 36-target cross-lab cohort which does not include four of the five pathological targets (Q96SW2, P15170, Q9Y2I7, P33981), so the variance-share argument holds at both the canonical 0.668 and the implied clean-cohort ceiling near 0.693.

\section{What Does Not Work}
\label{app:negative}

\subsection{PROTAC-STAN single-seed-to-multi-seed regression}
\label{app:negative:stan}

STAN ternary attention with single-seed evaluation reaches 0.718 but regresses to 0.656 under 3-seed validation; the attention mechanism contributes the $+0.05$ gain regardless of PLM input (zero-POI 0.714, random-POI 0.712, real-POI 0.718), indicating that the apparent improvement reflects the attention architecture rather than the protein representation. The single-seed-to-multi-seed regression of 0.062 AUROC parallels the 0.161 AUROC regression observed in the HPO V2 sweep (Section~\ref{sec:collapse:hpo}, Appendix~\ref{app:hpo}) and is consistent with the Bailey-L\'opez de Prado selection-bias mechanism applied to single-seed published-architecture evaluation; the gap between published 0.8833 random-CV and the 0.668 LOTO ceiling therefore decomposes into a target-overlap component (Ribes et al. 80 percent shared-target diagnostic), an attention-architecture-rather-than-PLM-representation component (the 0.05 gain regardless of POI input), and a single-seed selection-bias component (the 0.062 multi-seed regression).

\subsection{Other negative results}
\label{app:negative:other}

A representative sample of additional negative results is reported. SimCLR contrastive pre-training over Morgan fingerprint pairs (radius 2 versus radius 3, 128-dimensional embedding) yields $+0.012$ AUROC over the Morgan baseline under LOTO. Pseudo-labelling at confidence threshold 0.9 yields $+0.015$ AUROC under LOTO. Selective prediction filtering at top 10 percent confidence degrades AUROC from 0.659 to 0.621 at 10-seed canonical evaluation, an instance of the high-confidence inversion pattern documented in Section~\ref{sec:discussion} and Appendix~\ref{app:calibration}. Fragment-only Morgan fingerprints over warhead, linker, and E3 substructures all reduce LOTO AUROC by $-0.013$ to $-0.022$ relative to whole-molecule Morgan. IFP54 interaction fingerprints from docking poses reach 0.615 LOTO AUROC, $-0.023$ below the Morgan baseline. Boltz-2 ternary structure features alone reach 0.595 AUROC, statistically indistinguishable from chance; combined with the 2D pipeline they reach 0.664 ($\Delta = -0.002$, $p = 0.57$ versus the Morgan + warhead + ADMET baseline). Full structure-ladder detail in Appendix~\ref{app:geometric}. ProtoNet with Tanimoto distance reaches 0.679 at $k{=}5$, below RF retraining at $k{=}5$ (0.743). Pairwise continuous-DC50 ranking peaks at 0.655 at $k{=}25$, below the inter-laboratory noise floor of 0.678 estimated from the within-target cross-lab analysis.

\section{Reproducibility}
\label{app:reproduce}
\begin{figure}[h]
  \centering
  \includegraphics[width=0.85\linewidth]{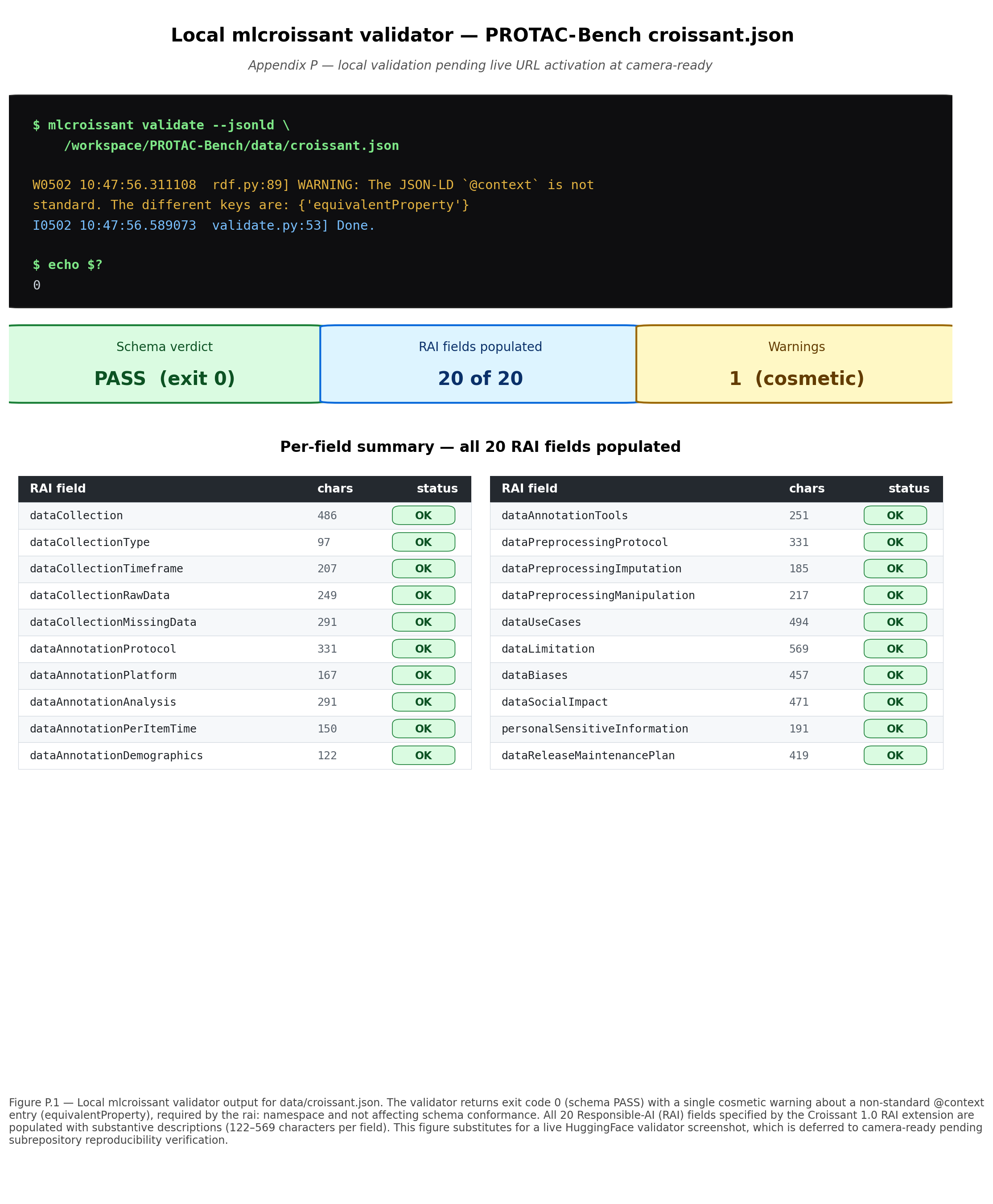}
  \caption{\textbf{Local mlcroissant validator output for the populated Croissant metadata.} The validator confirms schema PASS at exit code zero with all twenty MLCommons RAI extension fields populated; the only validator output is the cosmetic equivalentProperty warning shared with the upstream Croissant 1.0 schema. The validator report against the live HuggingFace URL is committed to the camera-ready release.}
  \label{fig:croissant_local}
\end{figure}

The released \texttt{reproduce.sh} script regenerates the canonical 
baseline, full-stack pipeline, and core robustness results from the 
canonical seed list \{7, 13, 29, 42, 43, 44, 53, 71, 89, 97\}. Total 
runtime is approximately 2 to 3 hours on a single CPU node with 16 
cores and 32 GB RAM. The full extended evaluation including HPO V2, 
EGNN comparison, and metadata enrichment requires GPU access (single 
RTX 4090 sufficient) and approximately 120 GPU-hours total. All seeds, hyperparameters, fingerprint configurations, and random states are documented in the code repository at \url{https://github.com/ThorKlm/PROTAC-Bench}. The Croissant metadata is at MLCommons Croissant 1.0 schema compliance with all twenty MLCommons RAI extension fields populated; local validation against the mlcroissant validator passes at exit code zero with only the cosmetic equivalentProperty warning shared with the upstream Croissant 1.0 schema (Figure~\ref{fig:croissant_local}, validator output reported in the released code repository). The dataset is hosted at \url{https://huggingface.co/datasets/ThorKl/protac-bench}; The validator report against the live HuggingFace URL is committed to a future revision.

\section{Ethics Statement}
\label{app:ethics}

\paragraph{Data sources and licensing.}
PROTAC-Bench is constructed from the publicly available PROTAC-DB and PROTAC-Pedia corpora curated by \citet{ribes2024modeling}, with the underlying PROTAC-Degradation-Predictor repository released under MIT License at \url{https://github.com/ribesstefano/PROTAC-Degradation-Predictor}; PROTAC-DB 3.0~\cite{ge2024protacdb} is published under CC BY 4.0. The merged 
dataset, fold assignments, and accompanying evaluation code are 
released under CC-BY-4.0 (dataset) and MIT (code). All compound 
activity measurements are derived from peer-reviewed publications 
and patent literature; no proprietary or restricted-access data are 
included. No human-subjects information is associated with any 
measurement; all data describe compound-target activity in cell-line 
or biochemical assays.

\paragraph{Dual-use considerations.}
PROTAC degraders are an emerging therapeutic modality with potential 
applications across oncology, immunology, and neurodegeneration. The 
benchmark and predictors released here are intended to support 
medicinal-chemistry enrichment of candidate libraries in academic and industrial drug-discovery 
settings. The current state of the field, as documented in this work, 
constrains practical claims to active-versus-inactive enrichment rather 
than to fine-grained potency ranking; predictors should not be 
deployed for selection of clinical candidates without further 
experimental validation. The architecture-invariant ceiling at the 
inter-laboratory reproducibility floor implies that no released model 
in this work exceeds the predictive accuracy achievable by domain 
experts using standard cheminformatics tools. The released benchmark 
is therefore methodologically aligned with reproducibility and 
evaluation transparency rather than with the production of novel 
predictive capabilities.

\paragraph{Broader impacts.}
The methodological pattern documented in this work, namely the 
decomposition of apparent generalisation gaps into measurement-variance 
components rather than learning-failure components, generalises in 
principle to other small-data therapeutic ML settings where 
inter-laboratory or inter-site variance is a dominant source of 
labelling noise. The cross-domain references in this work 
(drug-target interaction prediction~\cite{pahikkala2015toward}, 
pharmacogenomics~\cite{haibe2013inconsistency}, ecological habitat 
modelling~\cite{roberts2017cross}) carry similar variance structure. 
The benchmark and decomposition framework presented here provide a 
template for similar analyses in other domains, contributing to the 
broader trajectory of evaluation-protocol rigour in 
machine-learning-for-science settings.

\paragraph{Compute and environmental cost.}
All experiments reported in this work were conducted on a single 
Vast.ai instance with 2$\times$~NVIDIA GeForce RTX 4090 GPUs 
(450~W TBP each), 48 AMD Ryzen Threadripper CPU cores, and 64~GB RAM, 
over approximately three weeks of continuous utilisation. Total compute is approximately 1000 wall-clock hours of total GPU utilisation across the two RTX 4090s combined, including exploratory experiments, failed runs, the architecture-invariance comparison across eight architectures, hyperparameter optimisation across 2,000 trials, the metadata-enrichment pipeline, and the canonical evaluation runs reported in the main text. The corresponding energy consumption is approximately 450~kWh under typical RTX 4090 utilisation profiles, and the estimated carbon footprint is approximately 170~kg CO\textsubscript{2}eq under German grid intensity of 380~g CO\textsubscript{2}eq/kWh~\cite{lacoste2019quantifying}.
All compute, cloud infrastructure, and API costs were privately funded by the first author; no institutional, governmental, or commercial funding was received. Total project cost is on the order of a few hundred USD, and was fully privately funded, comprising Vast.ai GPU rental at on-demand pricing plus under 10 USD of LLM API spend for the metadata-enrichment pipeline (Appendix~\ref{app:metadata}). The released 
\texttt{reproduce.sh} script regenerates canonical baseline and 
full-stack results within approximately 2-3 hours on a single CPU 
node; full extended evaluation including the HPO V2 sweep and the 
EGNN comparison requires approximately 120 GPU-hours on a single 
RTX 4090, roughly one order of magnitude less than total project 
compute.

\paragraph{Reproducibility and access.}
All canonical experiments are reproducible from the released seed list 
and fold assignments. The \texttt{reproduce.sh} script regenerates the 
baseline and full-stack results on a single CPU node within 
approximately 2--3 hours. The released artefacts comprise the dataset 
CSV, fold assignment files, evaluation code, metadata-enrichment 
pipeline code, and Croissant metadata, hosted on HuggingFace under 
CC-BY-4.0 (dataset) and MIT (code).

\paragraph{Limitations on clinical translation.}
The benchmark and predictors presented here are research tools 
appropriate for medicinal-chemistry enrichment of compound libraries, not for clinical 
decision-making. The 0.668 LOTO ceiling under measurement-variance 
constraints, the 0.150 Expected Calibration Error documented in 
Appendix~\ref{app:calibration}, and the 0.19 per-target standard 
deviation collectively imply that predictions on individual compounds 
should not be used as substitutes for experimental validation in any 
clinical or regulatory context. The constructive contribution of this 
work is methodological (the decomposition of evaluation-protocol gaps 
into measurement-variance components and the demonstration of few-shot 
calibration as the locally-available recovery mechanism); the absolute 
predictive accuracy reported here remains constrained by 
inter-laboratory measurement variance and is not sufficient for 
deployment without prospective experimental validation.


\end{document}